\documentclass{article} % For LaTeX2e
\usepackage{iclr2027_conference,times}

\usepackage{amsmath,amsfonts,bm}

\def\eqref#1{equation~\ref{#1}}
\def\1{\bm{1}}

\DeclareMathAlphabet{\mathsfit}{\encodingdefault}{\sfdefault}{m}{sl}
\SetMathAlphabet{\mathsfit}{bold}{\encodingdefault}{\sfdefault}{bx}{n}

\usepackage{hyperref}
\usepackage{url}
\usepackage{graphicx}
\PassOptionsToPackage{numbers,sort&compress}{natbib}

\usepackage[utf8]{inputenc}
\usepackage[T1]{fontenc}

\usepackage{url}
\usepackage{graphicx}
\usepackage{nicefrac}

\usepackage{amsmath}
\usepackage{amssymb}
\usepackage{bm}

\usepackage{booktabs}
\usepackage{multirow}
\usepackage{array}
\usepackage{threeparttable}
\usepackage[table]{xcolor}
\usepackage{enumitem}
\usepackage[percent]{overpic}
\usepackage{pifont}
\usepackage{float}
\usepackage{subcaption}
\usepackage{wrapfig}
\usepackage{enumitem}
\usepackage{CJKutf8}

\usepackage{tabularx}
\usepackage{algorithm}
\usepackage{algpseudocode}

\usepackage{titletoc}
\usepackage{adjustbox}

\title{SignRAG: Unified Retrieval-Augmented \\ Gloss-Free Sign Language Translation}
\author{
\textbf{Zhi Rao}\textsuperscript{1},
\textbf{Yucheng Zhou}\textsuperscript{2},
\textbf{Qianran Sun}\textsuperscript{3},
\textbf{Yiqing Huang}\textsuperscript{1},
\textbf{Longcan Yuan}\textsuperscript{1},
\textbf{Jiayi Hou}\textsuperscript{4}, \\
\textbf{Chengwen Yao}\textsuperscript{5},
\textbf{Lin Cheng}\textsuperscript{5},
\textbf{Donghui Sun}\textsuperscript{5},
\textbf{Xiaoxin Chen}\textsuperscript{5},
\textbf{Jun Wan}\textsuperscript{1,3}\thanks{Corresponding author.}
\\[2mm]
\textsuperscript{1}Faculty of Innovation Engineering, Macau University of Science and Technology, Macau, China \\
\textsuperscript{2}SKL-IOTSC, CIS, University of Macau, Macau, China \\
\textsuperscript{3}MAIS, Institute of Automation, Chinese Academy of Sciences, Beijing, China \\
\textsuperscript{4}Yale University, New Haven, CT, USA \\
\textsuperscript{5}VIVO AI Lab, China \\
\texttt{zhir@student.must.edu.mo}
\qquad
\texttt{jun.wan@ia.ac.cn}
}

\iclrfinalcopy % Uncomment for camera-ready version, but NOT for submission.
\begin{document}

\maketitle

\begin{abstract}
Contemporary decoder-only large language models (LLMs) have demonstrated strong capabilities across a wide range of domains. However, existing pretraining paradigms for gloss-free sign language translation (SLT) are largely designed around conventional encoder-decoder pretrained language models, which limits their direct applicability to decoder-only LLMs. To address this limitation, we propose SignRAG, a unified framework combining hierarchical pretraining, target-domain retrieval augmentation, and retrieval-aware reinforcement fine-tuning. Hierarchical pretraining first learns linguistically grounded sign representations and then jointly aligns the sign encoder with an LLM, mitigating cross-modal optimization imbalance. For downstream adaptation, SignRAG complements parameter-based fine-tuning with a target-domain retrieval gallery that provides instance-specific translation cues. To ensure that retrieved contexts are used appropriately, we further introduce Retrieval Utility-Guided Reinforcement Fine-Tuning (RUG-RFT), which combines translation-quality and retrieval-utility rewards to encourage beneficial retrieval use while suppressing harmful reliance. Experiments on multiple SLT benchmarks establish new state-of-the-art performance. In particular, to the best of our knowledge, SignRAG is the first gloss-free approach to outperform gloss-supervised methods across all reported metrics on CSL-Daily. 
% Our code and models of different sizes will be publicly released.
% Our code has been released at \url{https://github.com/shahelaojieraozhi/SignRAG}, together with models of different sizes to support future academic research.
Our code has been released at \href{https://github.com/shahelaojieraozhi/SignRAG}{GitHub}, together with models of different sizes to support future academic research.
% Our code and pretrained models will be released after acceptance.
%Our code will be released at \url{xxx}, together with models of different sizes to support future academic research.
\end{abstract}

\section{Introduction}

Sign language translation (SLT) aims to translate visual sign sequences into natural language. Despite recent progress, existing methods still struggle to generalize across datasets with different signers, signing styles, and vocabulary distributions. A natural solution is to pretrain on large-scale sign language corpora and subsequently adapt the model to downstream SLT benchmarks.

Existing pretraining paradigms for gloss-free SLT can be broadly divided into two categories. \emph{Encoder-only pretraining} independently pretrains the sign encoder before transferring it to downstream translation models~\citep{GFSLT-VLP,Fla-LLM,SIGN2GPT,C2RL,VAP}, whereas \emph{end-to-end pretraining} jointly optimizes visual and language components on large-scale sign-language corpora~\citep{YT-ASL-SLT,zhang2024scaling,YouTube-SL-25,Uni-Sign,fish2026geo}. However, most existing methods still rely on T5-based encoder--decoder models, whose language modeling capacity remains limited compared with contemporary LLMs.

Replacing these decoders with a substantially stronger LLM is not a drop-in upgrade. Because the language branch is already strongly pretrained, it can dominate optimization and leave the comparatively weaker visual pathway under-trained. We refer to this phenomenon as \textbf{Cross-Modal Optimization Imbalance}. Such imbalance has been widely observed in multimodal learning~\citep{wang2020makes,huang2022modality,peng2022balanced} and can further induce shortcut learning driven by strong language priors~\citep{goyal2017making}. It raises a key question: \emph{how should large-scale sign pretraining be organized so that visual pathway still learns semantically grounded representations?}

\begin{figure}[!t]
\centering
\includegraphics[width=0.8\linewidth]{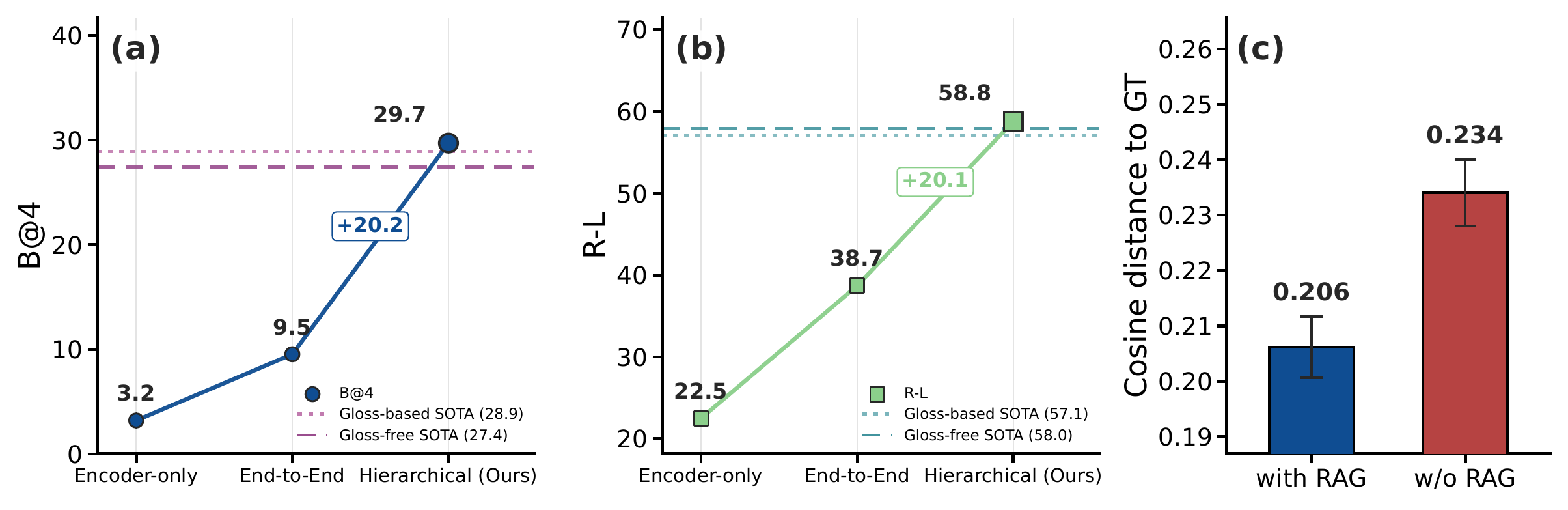}
\vspace{-3.5mm}
\caption{
\small
\textbf{Motivation.}
Hierarchical pretraining and retrieval augmentation improve sign-to-language learning.
\textbf{(a)}, \textbf{(b)} Comparison of different pretraining strategies. The solid and dashed lines denote the previous state-of-the-art performance under the gloss-free and gloss-based settings, respectively, corresponding to Geo-Sign~\citep{fish2026geo} and CV-SLT~\citep{CV-SLT}.
\textbf{(c)} Comparison of the semantic distance between generated and ground-truth (GT) texts with and without RAG. Semantic distance is measured using the independent sentence encoder \textit{paraphrase-multilingual-MiniLM-L12-v2}. Bars report the mean cosine distance over all samples, with error bars denoting $\pm$SEM. Samples are drawn from the CSL-Daily dev set.
}
\label{fig:intro}
\vspace{-5.5mm}
\end{figure}

We therefore propose a \textbf{Hierarchical Pretraining} strategy for LLM-based SLT. Inspired by the staged-alignment paradigm of general-purpose multimodal large language models (MLLMs)~\citep{llava,MPLUG-OwI2,Qwen2.5-VL,mm1}, we organize large-scale pretraining into two stages. We first use fine-grained sign--text supervision to learn linguistically grounded sign representations, and then jointly pretrain the sign encoder, multimodal projector, and LLM on the same corpus. In this way, the model first learns \emph{what the sign means} and then learns \emph{how to express its semantics in natural language}. As shown in Fig.~\ref{fig:intro}(a) and (b), hierarchical pretraining clearly outperforms alternative pretraining schemes, while the \texttt{Pretrain$\rightarrow$SFT} variant alone already surpasses previous state-of-the-art methods under both gloss-free and gloss-based settings.

After large-scale pretraining, downstream adaptation introduces a second challenge, which we term the \textbf{Target-Domain Adaptation Dilemma}. Sign-language datasets exhibit substantial distribution shifts across languages, recording conditions, and signers, while downstream training sets are typically much smaller than the pretraining corpus. Aggressive fine-tuning may therefore erode transferable knowledge acquired during pretraining, whereas conservative fine-tuning may leave the target domain insufficiently adapted. To address this dilemma, we complement parametric SFT with a \textbf{non-parametric target-domain semantic memory}. Specifically, we construct a retrieval gallery exclusively from the target-domain training split and retrieve semantically related sign instances together with their translations as instance-specific linguistic cues. As shown in Fig.~\ref{fig:intro}(c), retrieval augmentation reduces the semantic distance between generated and GT texts from $0.234$ to $0.206$.

Retrieval, however, is not always beneficial, leading to a third challenge that we term the \textbf{Retrieval Reliance Dilemma}. Standard SFT optimizes token-level likelihood but provides no explicit supervision on whether retrieved contexts actually benefit generation, making both underutilization and harmful reliance difficult to regulate. We first introduce retrieval dropout during retrieval-augmented SFT to improve robustness when retrieved information is absent. We then propose \textbf{Retrieval Utility-Guided Reinforcement Fine-Tuning (RUG-RFT)} based on Group Relative Policy Optimization (GRPO)~\citep{shao2024deepseekmath}. RUG-RFT combines translation-quality and retrieval-utility rewards, encouraging the model to exploit informative retrieval while suppressing harmful reliance on misleading contexts.

We conduct extensive experiments on four downstream SLT benchmarks: CSL-Daily, PHOENIX-2014T, How2Sign, and OpenASL. SignRAG consistently outperforms existing gloss-free state-of-the-art methods. On CSL-Daily, it further surpasses gloss-supervised methods across all metrics; to the best of our knowledge, this is the first gloss-free SLT approach to achieve such performance. Our main contributions are summarized as follows:
\begin{itemize}[leftmargin=*, itemsep=-2pt, topsep=-2pt]
    \item We propose a hierarchical pretraining paradigm that first learns linguistically grounded sign representations through fine-grained sign--text alignment and then performs large-scale joint multimodal pretraining with a contemporary LLM.
    
    \item We introduce a target-domain retrieval-augmented adaptation mechanism that constructs a non-parametric semantic memory to complement parameter-based SFT with instance-specific target-domain knowledge.
    
    \item We propose Retrieval Utility-Guided Reinforcement Fine-Tuning (RUG-RFT), which jointly optimizes translation quality and retrieval contribution to enable more effective and robust use of retrieved contexts.
    
    \item Extensive experiments across multiple SLT benchmarks validate the effectiveness of the proposed framework and establish new state-of-the-art performance.
\end{itemize}

%Experiments on four SLT benchmarks show strong gloss-free translation performance without external large-scale sign-language pre-training, with ablations confirming the effectiveness of ARCS and RUG-RFT.
% 新的预训练方式.
% Retrieval Dropout Augmentation (RDA)
% 参考一下，写一下 RAG
% 本文从伪 Gloss 生成角度探索基于伪标签的预训练策略：PG 一旦生成，便可以作为高质量、可重复使用的资产，从而避免 SLT 模型重复采用复杂预训练设计（如 VAP）。因此，后续训练流程可以更简单、更高效，同时仍保持有竞争力的性能和良好可扩展性。
% 如果质疑实际落地: 设置实验。

\section{Related Work}
\label{sec:related_work}

\textbf{Sign language translation.} Early SLT relies on gloss supervision via gloss recognition or cascaded translation~\citep{SLRT,SignBT,TS-SLT,MMTLB}, whereas gloss-free SLT maps sign videos directly to text using vision--language learning, LLM decoding, or cross-modal alignment~\citep{NSLT,TSPNet,CSGCR,GASLT,GFSLT-VLP,SIGN2GPT,Fla-LLM,Sign-LLM,VAP,C2RL}, and external large-scale pretraining further improves transferability~\citep{YT-ASL-SLT,SSVP-SLT,Uni-Sign}. \textbf{Retrieval-augmented generation and reinforcement learning.} RAG conditions generation on retrieved non-parametric contexts~\citep{lewis2020retrieval,chen2022murag}, retrieval-based SLT~\citep{RASLT} mainly targets gloss-based settings, and GRPO~\citep{shao2024deepseekmath,guo2025deepseek} has been applied to retrieval-enhanced reasoning and generation~\citep{huang2025rag,li2025r3,wang2025vrag}. SignRAG studies \emph{how} large-scale sign pretraining should be organized under a modern LLM decoder, and complements parametric adaptation with a sign-to-text retrieval gallery and retrieval Utility-Guided reinforcement fine-tuning, without gloss annotations. A more detailed discussion is provided in Appendix~\ref{app:related_work}.

\section{Method}\label{sec:method}

\begin{figure*}[!t]
\centering
\includegraphics[width=\linewidth]{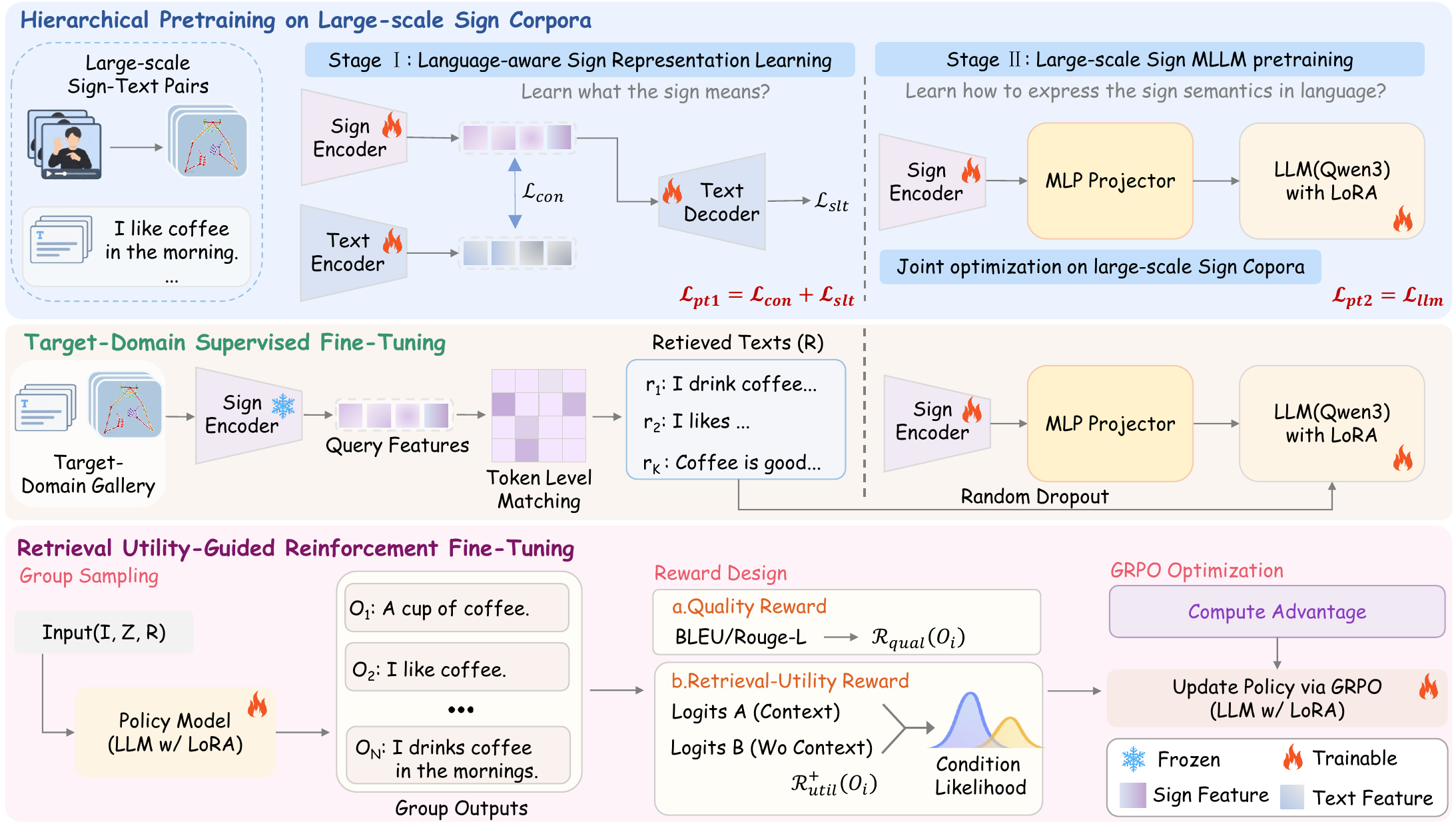}
\vspace{-3mm}
\caption{
\small
\textbf{Overview of SignRAG.}
\textbf{(1)} Hierarchical pretraining on large-scale sign corpora: Stage~I learns language-aware sign representations with a contrastive objective $\mathcal{L}_{\mathrm{con}}$ and a skeleton-to-text objective $\mathcal{L}_{\mathrm{slt}}$; Stage~II couples the pretrained sign encoder with the LLM (Qwen3) via a multimodal projector and performs large-scale joint pretraining.
\textbf{(2)} Target-domain supervised fine-tuning: a \emph{sign retrieval gallery} provides semantically related reference translations as auxiliary cues, and the model is fine-tuned with sign features and retrieved texts, with random retrieval dropout.
\textbf{(3)} Retrieval utility-guided reinforcement fine-tuning: a quality reward and a retrieval-utility reward are combined and optimized with GRPO.
}
\label{fig:method}
\vspace{-3mm}
\end{figure*}

As illustrated in Fig.~\ref{fig:method}, \textbf{SignRAG} consists of three stages for gloss-free sign language translation (SLT). \textbf{Stage 1} performs hierarchical pretraining, where Stage~I learns linguistically grounded sign representations and Stage~II jointly pretrains the sign encoder, multimodal projector, and LLM. \textbf{Stage 2} performs retrieval-augmented SFT using a target-domain \emph{Sign Retrieval Gallery}, which retrieves semantically related sign instances and provides their paired translations as auxiliary cues; retrieval dropout is further applied during training. \textbf{Stage 3} applies \textbf{Retrieval Utility-Guided Reinforcement Fine-Tuning (RUG-RFT)} with GRPO, combining a Quality Reward and a Retrieval-Utility Reward to improve translation quality and encourage appropriate use of retrieved contexts. We next describe these three components in detail.

\subsection{Hierarchical Pretraining}\label{sec:pre-training}

\paragraph{Stage I: Sign--text alignment.}
We first align skeletal sign representations with natural-language semantics.
Given a sign video $V$, MMPose~\citep{mmpose} first extracts 2D keypoints, which are subsequently encoded by a CoSign-based skeleton encoder~\citep{cosign} into frame-level skeletal representations.
The paired translation $T$ is encoded by an mBART-based text encoder~\citep{mbart}.
The two encoders are jointly optimized with a bidirectional InfoNCE contrastive objective $\mathcal{L}_{\mathrm{con}}$~\citep{cico} and a CoCa-style skeleton-to-text translation objective $\mathcal{L}_{\mathrm{slt}}$~\citep{coca}:
\begin{equation}
    \mathcal{L}_{\mathrm{pt}}
    =
    \mathcal{L}_{\mathrm{con}}
    +
    \mathcal{L}_{\mathrm{slt}}.
\end{equation}
Detailed formulations are provided in Appendix~\ref{app:pretrain}.
This stage endows the sign encoder with linguistically grounded representations before it is coupled with the LLM.

\paragraph{Stage II: Large-scale sign MLLM pretraining.}
We then couple the aligned sign encoder with an LLM. A GELU-activated MLP projector maps the skeletal features into the embedding space of Qwen3~\citep{qwen3}, and LoRA adapters are attached to the query and value projections of the LLM self-attention layers. The sign encoder, the projector, and the LoRA-adapted LLM are jointly optimized on large-scale sign--text corpora, CSL-News~\citep{Uni-Sign} for Chinese sign language and YouTube-ASL~\citep{YT-ASL-SLT} for American sign language, with the autoregressive objective
\begin{equation}
    \mathcal{L}_{\mathrm{lm}}
    =
    -\sum_{t=1}^{T}
    \log p_\theta\!\left(y_t \mid y_{<t}, \mathbf{I}, \mathbf{Z}\right),
\end{equation}
where $\mathbf{Z}$ denotes the projected sign features and $\mathbf{I}$ the instruction.
Because the sign encoder enters joint training already grounded in language semantics, it no longer acts as the weak branch when coupled with the LLM, which mitigates the Cross-Modal Optimization Imbalance discussed in the Introduction.

\subsection{Retrieval-Augmented SFT}\label{sec:SFT}

\paragraph{Sign retrieval gallery.}
For each downstream dataset, we fine-tune the sign encoder on its training set and build a \emph{sign-to-text} retrieval gallery from that split only; validation and test samples serve exclusively as queries, so that retrieved knowledge always comes from the target domain.
For a gallery video $V^{g}$, its frame-level representation is
\begin{equation}
    Z^{g}
    =
    \texttt{SkeletonEnc}(V^{g})
    \in
    \mathbb{R}^{L_g \times C},
\end{equation}
where $L_g$ denotes the temporal length and $C$ is the feature dimension.
Each gallery entry stores the encoded sign feature sequence and its paired ground-truth translation,
\begin{equation}
    \mathcal{G}_n
    =
    \left(
    Z^{g}_n,
    T^{g}_n
    \right).
\end{equation}
Query videos are encoded by the same skeleton encoder for sign-level retrieval.

\paragraph{Attention-weighted sign--sign similarity.}
Given a query sign sequence
$Z^{q}\in\mathbb{R}^{L_q\times C}$
and a gallery sequence
$Z^{g}\in\mathbb{R}^{L_g\times C}$,
we first project their frame-level features into a shared $D$-dimensional embedding space:
\begin{equation}
    Q
    =
    \operatorname{Norm}\!\left(\phi_q(Z^{q})\right),
    \qquad
    G
    =
    \operatorname{Norm}\!\left(\phi_g(Z^{g})\right),
\end{equation}
where
$Q\in\mathbb{R}^{L_q\times D}$ and
$G\in\mathbb{R}^{L_g\times D}$.
Here, $\phi_q(\cdot)$ and $\phi_g(\cdot)$ denote the projection heads, and
$\operatorname{Norm}(\cdot)$ performs row-wise $\ell_2$ normalization.

We then compute the frame-level cosine similarity between the query and gallery sequences:
\begin{equation}
    E_{i,j}
    =
    Q_i^{\top}G_j,
\end{equation}
where $E\in\mathbb{R}^{L_q\times L_g}$.
For each query frame, we normalize its similarities over the gallery sequence to obtain attention weights:
\begin{equation}
    P_{i,j}
    =
    \frac{\exp(E_{i,j})}
    {\sum_{j'=1}^{L_g}\exp(E_{i,j'})}.
\end{equation}
The global sign--sign similarity is obtained by aggregating the attention-weighted frame-level similarities:
\begin{equation}
\label{eq:sign_sign_sim}
    S(Q,G)
    =
    \frac{1}{L_q}
    \sum_{i=1}^{L_q}
    \sum_{j=1}^{L_g}
    P_{i,j}E_{i,j}.
\end{equation}

Compared with pooling an entire sign video into a single global representation, our formulation preserves fine-grained frame-level correspondences between sign sequences, which are crucial for retrieving semantically relevant samples. For clarity, the padding masks used to handle variable-length sequences are omitted from the above formulations. To improve memory efficiency, we shard the gallery across distributed workers and compute sign--sign similarities in query--gallery chunks. The exact global top-$K$ results are then obtained by merging the local top-$K$ candidates returned by all workers. Further implementation details are provided in Appendix~\ref{app:retrieval_detail}.

\paragraph{Retrieval-augmented fine-tuning.}
We use skeleton embeddings $Z_s$ as the visual representation of sign language and project them into the LLM latent space:
\begin{equation}
Z = \mathrm{MLP}(Z_s),
\end{equation}
where the projector is a GELU-activated MLP.
We use Qwen3~\citep{qwen3} as the base LLM and form its input by concatenating the instruction $\mathbf{I}$, sign features $\mathbf{Z}$, and retrieved contexts $\mathbf{R}$.

To support the subsequent Retrieval-Utility Reward, we apply retrieval dropout during SFT:
\begin{equation}
\tilde{\mathbf{R}} =
\begin{cases}
\mathbf{R}, & \text{with probability } 1-p, \\
\varnothing, & \text{with probability } p,
\end{cases}
\end{equation}
where $\varnothing$ denotes an empty token sequence without placeholder tokens.
The model is fine-tuned with LoRA by minimizing
\begin{equation}
\mathcal{L}_{\mathrm{sft}}
= -\sum_{t=1}^{T} \log p_\theta\!\left(y_t \mid y_{<t}, \mathbf{I}, \mathbf{Z}, \tilde{\mathbf{R}}\right),
\end{equation}
where $T$ is the target sequence length and $\theta$ denotes the trainable LoRA parameters.
This encourages coherent generation under both retrieval-available and retrieval-absent conditions.

\subsection{Retrieval Utility-Guided Reinforcement Fine-Tuning}\label{sec:RFT}

SFT optimizes token-level likelihood and provides no explicit signal for whether retrieved contexts are useful.
We therefore introduce RUG-RFT to optimize sentence-level quality while regulating retrieval reliance.
RUG-RFT combines a \emph{Quality Reward} with a \emph{Retrieval-Utility Reward} under Group Relative Policy Optimization (GRPO)~\citep{shao2024deepseekmath}, which estimates advantages from intra-group reward statistics without a separate value model.

\paragraph{GRPO with group sampling.}
Given an input $(\mathbf{I}, \mathbf{Z}, \mathbf{R})$, the current policy $\pi_{\theta}$ samples $N$ candidate outputs $\{O_1,\ldots,O_N\}$, each scored by $\mathcal{R}(O_i)$ to obtain $r_i$.
GRPO standardizes rewards within the group to compute
\begin{equation}
A_i = \frac{r_i - \text{mean}\{r_1, \ldots, r_N\}}{\text{std}\{r_1, \ldots, r_N\}}.
\label{eq:advantage}
\end{equation}
Let $c_1=\pi_{\theta}(O_i\mid\mathbf{I},\mathbf{Z},\mathbf{R})/\pi_{\text{old}}(O_i\mid\mathbf{I},\mathbf{Z},\mathbf{R})$ and $c_2=\text{clip}(c_1,1-\varepsilon,1+\varepsilon)$.
The policy is updated by maximizing
\begin{equation}
\mathcal{J}_{\text{GRPO}}(\theta) = \mathbb{E}\!\left[\frac{1}{N}\sum_{i=1}^{N} \min(c_1 A_i,\, c_2 A_i) - \beta\, D_{\text{KL}}(\pi_{\theta}\|\pi_{\text{ref}})\right],
\label{eq:grpo_main}
\end{equation}
where the KL term regularizes deviation from the reference policy $\pi_{\text{ref}}$.

\subsubsection{Retrieval-Utility Reward}

We quantify the contribution of retrieval for a \emph{fixed output} using teacher-forcing conditional likelihoods.
Given $(\mathbf{I},\mathbf{Z},\mathbf{R})$, the current policy $\pi_\theta$ generates $Y=(y_1,\ldots,y_T)$ under the retrieval-augmented condition.
We then score the same $Y$ with retrieval using $\pi_\theta$ and without retrieval using a frozen reference policy $\pi_{\text{ref}}$:
\begin{align}
\ell^{\text{with}}(Y)
&=
\frac{1}{T}
\sum_{t=1}^{T}
\log p_\theta
\!\left(
y_t
\mid
y_{<t},
\mathbf{I},
\mathbf{Z},
\mathbf{R}
\right),
\\
\ell^{\text{wo}}(Y)
&=
\frac{1}{T}
\sum_{t=1}^{T}
\log p_{\text{ref}}
\!\left(
y_t
\mid
y_{<t},
\mathbf{I},
\mathbf{Z}
\right).
\end{align}
Here, $\pi_{\text{ref}}$ is the frozen SFT checkpoint and is identical to the KL anchor in Eq.~\ref{eq:grpo_main}.
Using $\pi_{\text{ref}}$ for $\ell^{\text{wo}}(Y)$ prevents the policy from increasing the reward by artificially lowering its retrieval-free likelihood.
We define the retrieval-induced likelihood gain as
\begin{equation}
\Delta \ell_{\mathrm{ret}}(Y)
=
\ell^{\text{with}}(Y)
-
\ell^{\text{wo}}(Y),
\end{equation}
where a positive value indicates that retrieval increases the model's confidence in $Y$.

\subsubsection{Quality Reward and Reward Composition}

To align with task metrics while regulating retrieval reliance, we combine a sentence-level Quality Reward with the Retrieval-Utility Reward.
Let $\mathcal{R}_{\text{qual}}(Y)$ denote a translation-quality score, such as BLEU or ROUGE, computed against the ground truth.
We retain only positive retrieval utility:
\begin{equation}
\mathcal{R}_{\text{util}}^{+}(Y)
=
\max\!\left(
0,\,
\Delta \ell_{\mathrm{ret}}(Y)
\right).
\end{equation}
To balance the magnitudes of the two rewards, we introduce
\begin{equation}
\gamma
=
\frac{
\widehat{\mathbb{E}}
\big[
|\mathcal{R}_{\text{qual}}|
\big]
}{
\max\!\left(
\widehat{\mathbb{E}}
\big[
\mathcal{R}_{\text{util}}^{+}
\big],
\,\tau
\right)
},
\end{equation}
where $\widehat{\mathbb{E}}[\cdot]$ is an exponential moving average over recent batches, $\tau$ prevents early instability, and $\gamma$ is treated as stop-gradient.
The total reward used in Eq.~\ref{eq:advantage} is
\begin{equation}
\mathcal{R}_{\text{total}}(Y)
=
\mathcal{R}_{\text{qual}}(Y)
+
\alpha \cdot \gamma \cdot
\mathcal{R}_{\text{util}}^{+}(Y),
\end{equation}
where $\alpha$ controls the strength of the Retrieval-Utility Reward.

\section{Experiment}
\subsection{Experimental Setup}

\paragraph{Datasets and metrics.}
We use CSL-News~\citep{Uni-Sign} and YouTube-ASL~\citep{YT-ASL-SLT} as large-scale pretraining datasets for Chinese Sign Language and American Sign Language, respectively. We then evaluate SignRAG on four downstream SLT benchmarks: CSL-Daily~\citep{SignBT}, PHOENIX-2014T~\citep{NSLT}, How2Sign~\citep{how2sign}, and OpenASL~\citep{OpenASL}. Their train/validation/test splits are 18,401/1,077/1,176, 7,096/519/642, 31,085/1,739/2,348, and 96,476/966/975, respectively. We report BLEU-1/2/3/4~\citep{bleu} and ROUGE-L~\citep{rouge} following standard SLT evaluation protocols.

\paragraph{Retrieval gallery protocol.}
For each dataset, we build a separate train-only retrieval gallery from its training split, where each entry stores the encoded sign representation and paired translation. Validation and test samples are used only as queries and are never included in the gallery; during training, the current query instance is excluded to prevent self-retrieval. The details are shown in Appendix~\ref{app:identical_retrieval}. A detailed discussion of the efficiency, computational cost, and retrieval scalability of SignRAG is provided in Appendix~\ref{sec:efficiency}.

\begin{table*}[t]
\scriptsize
\centering
\caption{
Main results on four SLT benchmarks.
``w/ gloss'' and ``w/o gloss'' denote methods with and without gloss supervision, respectively.
``Ext.'' indicates whether external sign-language datasets are used for pre-training.
For the ``w/o gloss'' setting, methods with and without external pre-training data are separated and compared independently.
Bold scores indicate the best results within each dataset, gloss setting, and external-data setting.
``Venue'' denotes the publication venue and year.
}
\label{tab:main_results_compact}

\begingroup
\setlength{\tabcolsep}{0.65pt}
\renewcommand{\arraystretch}{0.85}

\definecolor{markgreen}{RGB}{40,120,70}
\newcommand{\extyes}{{\color{markgreen}\ding{51}}}
\newcommand{\extno}{\ding{55}}

\begin{tabular*}{\textwidth}{@{}ccllc@{\extracolsep{\fill}}rrrrr@{}}
\toprule
\textbf{Dataset}
& \textbf{Gloss}
& \textbf{Method}
& \textbf{Venue}
& \textbf{Ext.}
& \textbf{R-L$\uparrow$}
& \textbf{B@1$\uparrow$}
& \textbf{B@2$\uparrow$}
& \textbf{B@3$\uparrow$}
& \textbf{B@4$\uparrow$}\\
\midrule

% =========================================================
% CSL-Daily
% =========================================================
\multirow{15}{*}{\rotatebox[origin=c]{90}{CSL-Daily}}
& \multirow{4}{*}{w/ gloss}
& SignBT~\citep{SignBT}
& CVPR'21
& \extno
& 49.31 & 51.42 & 37.26 & 27.76 & 21.34 \\
&
& SLTUNET~\citep{SLTUNET}
& ICLR'23
& \extno
& 54.08 & 54.98 & 41.44 & 31.84 & 25.01 \\
&
& TS-SLT~\citep{TS-SLT}
& NeurIPS'22
& \extno
& 55.72 & 55.44 & 42.59 & 32.87 & 25.79 \\
&
& CV-SLT~\citep{CV-SLT}
& AAAI'24
& \extno
& \textbf{57.06}
& \textbf{58.29}
& \textbf{45.15}
& \textbf{35.77}
& \textbf{28.94} \\

\cmidrule(lr){2-10}

& \multirow{11}{*}{w/o gloss}
& GFSLT-VLP~\citep{GFSLT-VLP}
& ICCV'23
& \extno
& 36.44 & 39.37 & 24.93 & 16.26 & 11.00 \\
&
& Sign2GPT~\citep{SIGN2GPT}
& ICLR'24
& \extno
& 42.36 & 41.75 & 28.73 & 20.60 & 15.40 \\
&
& VAP~\citep{VAP}
& ECCV'24
& \extno
& 48.56 & 49.99 & -- & -- & 20.85 \\
&
& C$^2$RL~\citep{C2RL}
& TCSVT'25
& \extno
& 48.21 & 49.32 & 36.28 & 27.54 & 21.61 \\
&
& RVLF~\citep{Rao_2026_CVPR}
& Findings CVPR'26
& \extno
& 55.92 & 54.96 & 42.45 & 33.33 & 26.71 \\
&
& \textbf{SignRAG-SFT}
& --
& \extno
& 55.43 & 54.22 & 41.07 & 32.21 & 26.06 \\
&
& \textbf{SignRAG-RFT}
& --
& \extno
& \textbf{56.36}
& \textbf{56.60}
& \textbf{43.87}
& \textbf{34.43}
& \textbf{27.73} \\

\cmidrule(lr){3-10}

&
& Uni-Sign~\citep{Uni-Sign}
& ICLR'25
& \extyes
& 56.51 & 55.08 & 42.14 & 32.98 & 26.36 \\
&
& Geo-Sign~\citep{fish2026geo}
& NeurIPS'25
& \extyes
& 57.97 & 55.89 & -- & -- & 27.42 \\
&
& \textbf{SignRAG-SFT}
& --
& \extyes
& 60.20 & 59.39 & 46.94 & 37.95 & 31.22 \\
&
& \textbf{SignRAG-RFT}
& --
& \extyes
& \textbf{62.34}
& \textbf{61.12}
& \textbf{48.74}
& \textbf{39.73}
& \textbf{32.91} \\

% =========================================================
% PHOENIX-2014T
% =========================================================
\midrule
\multirow{11}{*}{\rotatebox[origin=c]{90}{PHOENIX-2014T}}
& \multirow{3}{*}{w/ gloss}
& SignBT~\citep{SignBT}
& CVPR'21
& \extno
& 49.54 & 50.80 & 37.75 & 29.72 & 24.32 \\
&
& TS-SLT~\citep{TS-SLT}
& NeurIPS'22
& \extno
& 53.48 & \textbf{54.90} & 42.43 & 34.46 & 28.95 \\
&
& CV-SLT~\citep{CV-SLT}
& AAAI'24
& \extno
& \textbf{54.33}
& 54.88
& \textbf{42.68}
& \textbf{34.79}
& \textbf{29.27} \\

\cmidrule(lr){2-10}

& \multirow{8}{*}{w/o gloss}
& GFSLT-VLP~\citep{GFSLT-VLP}
& ICCV'23
& \extno
& 42.49 & 43.71 & 33.18 & 26.11 & 21.44 \\
&
& Sign2GPT~\citep{SIGN2GPT}
& ICLR'24
& \extno
& 48.90 & 49.54 & 35.96 & 28.83 & 22.52 \\
&
& FLa-LLM~\citep{Fla-LLM}
& LREC-COLING'24
& \extno
& 45.27 & 46.29 & 35.33 & 28.03 & 23.09 \\
&
& VAP~\citep{VAP}
& ECCV'24
& \extno
& 51.28 & 53.07 & -- & -- & 26.16 \\
&
& C$^2$RL~\citep{C2RL}
& TCSVT'25
& \extno
& 50.96 & 52.81 & 40.20 & 32.20 & 26.75 \\
&
& RVLF~\citep{Rao_2026_CVPR}
& Findings CVPR'26
& \extno
& 52.56 & 53.60 & 41.20 & 33.24 & 27.86 \\
&
& \textbf{SignRAG-SFT}
& --
& \extno
& 52.14 & 53.05 & 40.88 & 32.68 & 27.30 \\
&
& \textbf{SignRAG-RFT}
& --
& \extno
& \textbf{53.07}
& \textbf{54.09}
& \textbf{42.62}
& \textbf{34.31}
& \textbf{28.63} \\

% =========================================================
% How2Sign
% =========================================================
\midrule
\multirow{12}{*}{\rotatebox[origin=c]{90}{How2Sign}}
& \multirow{12}{*}{w/o gloss}
& GloFE-VN~\citep{GloFE-VN}
& ACL'23
& \extno
& 12.6 & 14.9 & 7.3 & 3.9 & 2.2 \\
&
& OpenSLT~\citep{OpenSLT}
& CVPRW/WiCV'23
& \extno
& -- & 34.0 & 19.3 & 12.2 & 8.0 \\
&
& C$^2$RL~\citep{C2RL}
& TCSVT'25
& \extno
& 27.0 & 29.1 & 18.6 & 12.9 & 9.4 \\
&
& FLa-LLM~\citep{Fla-LLM}
& LREC-COLING'24
& \extno
& 27.8 & 29.8 & 19.0 & 13.3 & 9.7 \\
&
& VAP~\citep{VAP}
& ECCV'24
& \extno
& 27.8 & 39.2 & -- & -- & 12.9 \\
&
& RVLF~\citep{Rao_2026_CVPR}
& Findings CVPR'26
& \extno
& 34.7 & 39.4 & 25.3 & 18.6 & 14.3 \\
&
& \textbf{SignRAG-SFT}
& --
& \extno
& 34.7 & 40.2 & 25.9 & 18.4 & 13.7 \\
&
& \textbf{SignRAG-RFT}
& --
& \extno
& \textbf{36.2} & \textbf{42.9} & \textbf{28.0} & \textbf{20.3} & \textbf{15.2} \\

\cmidrule(lr){3-10}

&
& YouTube-ASL~\citep{YT-ASL-SLT}
& NeurIPS D\&B'23
& \extyes
& -- & 37.8 & 24.1 & 16.9 & 12.4 \\
&
& SignMusketeers~\citep{signmusketeers}
& Findings ACL'25
& \extyes
& -- & 41.5 & 27.2 & 19.3 & 14.3 \\
&
& Uni-Sign~\citep{Uni-Sign}
& ICLR'25
& \extyes
& 36.0 & 40.2 & 27.1 & 19.7 & 14.9 \\
&
& SSVP-SLT~\citep{SSVP-SLT}
& ACL'24
& \extyes
& 38.4 & 43.2 & 28.8 & 20.8 & 15.5 \\
&
& \textbf{SignRAG-SFT}
& --
& \extyes
& 36.7 & 43.3 & 28.6 & 20.7 & 15.6 \\
&
& \textbf{SignRAG-RFT}
& --
& \extyes
& \textbf{38.5} & \textbf{44.7} & \textbf{30.3} & \textbf{22.1} & \textbf{16.7} \\

% =========================================================
% OpenASL
% =========================================================
\midrule
\multirow{9}{*}{\rotatebox[origin=c]{90}{OpenASL}}
& \multirow{9}{*}{w/o gloss}
% & I3D-transformer~\citep{OpenASL}
% & EMNLP'22
% & \extno
% & 18.64 & 18.31 & 10.15 & 7.19 & 5.66 \\
% &
& OpenASL~\citep{OpenASL}
& EMNLP'22
& \extno
& 21.02 & 20.92 & 12.08 & 8.59 & 6.72 \\
&
& GloFE-VN~\citep{GloFE-VN}
& ACL'23
& \extno
& 21.75 & 21.56 & 12.74 & 9.05 & 7.06 \\
&
& C$^2$RL~\citep{C2RL}
& TCSVT'25
& \extno
& 31.36 & 31.46 & 21.85 & 16.58 & 13.21 \\
&
& VAP~\citep{VAP}
& ECCV'24
& \extno
& 41.38 & 45.92 & -- & -- & 21.23 \\
&
& RVLF~\citep{Rao_2026_CVPR}
& Findings CVPR'26
& \extno
& 43.92 & 48.73 & 35.40 & 27.99 & 22.84 \\
&
& \textbf{SignRAG-SFT}
& --
& \extno
& 42.96 & 46.24 & 34.25 & 27.12 & 22.24 \\
&
& \textbf{SignRAG-RFT}
& --
& \extno
& \textbf{43.87}
& \textbf{49.14}
& \textbf{36.58}
& \textbf{29.01}
& \textbf{23.61} \\

\cmidrule(lr){3-10}

&
& Uni-Sign~\citep{Uni-Sign}
& ICLR'25
& \extyes
& 43.22 & 49.35 & 36.22 & 28.55 & 23.14 \\
&
& \textbf{SignRAG-SFT}
& --
& \extyes
& 45.33 & 50.31 & 38.45 & 30.28 & 25.12 \\
&
& \textbf{SignRAG-RFT}
& --
& \extyes
& \textbf{46.91}
& \textbf{51.56}
& \textbf{39.52}
& \textbf{32.06}
& \textbf{26.62} \\

\bottomrule
\end{tabular*}
\endgroup

\vspace{-2mm}
\end{table*}

\paragraph{Implementation details.}
We adopt CoSign~\citep{cosign} with a three-layer
Transformer-based skeleton encoder and use Qwen3-8B as the
decoder-only LLM across all datasets. All experiments are conducted on four
NVIDIA RTX 4090D GPUs, each equipped with 48~GB of memory, using bfloat16
precision. The first and second pre-training stages are run for 200 and 50
epochs, respectively, followed by 10 epochs of SFT and 2 epochs of RFT. We use
SGD with a learning rate of $0.01$ in the first pre-training stage and AdamW in
all subsequent stages. The learning rates for SFT and RFT are set to $1 \times
10^{-4}$ and $1 \times 10^{-5}$, respectively, with a retrieval dropout
probability of $p=0.5$. Checkpoints are selected based on validation
BLEU-4 scores. For the number of retrieved items, we set $K=3$. Full
implementation details are provided in Appendix~\ref{sec:More Details}.

\subsection{Main Results}
Table~\ref{tab:main_results_compact} summarizes the main results. SignRAG outperforms existing state-of-the-art methods across all evaluated datasets. In particular, on CSL-Daily, SignRAG further surpasses gloss-based approaches, suggesting that large-scale pretraining on external sign language data, together with the capabilities of modern LLMs, can partially compensate for the absence of costly gloss annotations. 
Even without pretraining on additional data, SignRAG achieves state-of-the-art performance under the same setting. The training protocol follows RVLF~\citep{Rao_2026_CVPR}. It is worth noting that, due to the lack of a large-scale German sign language pretraining dataset, all results on PHOENIX-2014T are obtained without pretraining on additional data. Compared with the Chinese datasets, the gains of SignRAG over existing state-of-the-art methods are more moderate on the English datasets. This discrepancy may be attributed to differences in dataset difficulty and evaluation protocols between Chinese and English benchmarks. In addition, the YouTube-ASL data used for pretraining are incomplete, which may limit the effectiveness of large-scale pretraining on English sign language translation. Further details are provided in Appendix~\ref{app:dataset_statistics}. For a fair comparison, we do not include methods that use substantially larger pretraining datasets or introduce modules specifically designed for English SLT datasets~\citep{zhang2024scaling,jang2025lost}.

\subsection{Ablation Study}
Table~\ref{tab:main_ablation}(a) evaluates the contributions of pretraining, retrieval augmentation, and RUG-RFT on CSL-Daily. Pretraining improves ROUGE-L from 55.03 to 58.78 and BLEU-4 from 23.46 to 29.72. Adding retrieval further raises them to 60.20 and 31.22, respectively. With RUG-RFT, performance reaches 62.34 ROUGE-L and 32.91 BLEU-4, with consistent gains across all metrics. These results demonstrate the effectiveness of progressively introducing pretraining, retrieval, and retrieval utility-guided reinforcement fine-tuning. Table~\ref{tab:main_ablation}(b) analyzes the reward design. Starting from the same retrieval-augmented SFT model, both the Quality Reward and Retrieval-Utility Reward improve performance independently. Their combination achieves the best results across all metrics, reaching 62.34 ROUGE-L and 32.91 BLEU-4. This confirms that translation quality and retrieval utility provide complementary optimization signals for RUG-RFT.

\begin{table}[t]
\centering
\small
\caption{Ablation studies on CSL-Daily. Components in (a) are added cumulatively. Reward variants in (b) start from the same retrieval-augmented SFT model.}
\label{tab:main_ablation}
\setlength{\tabcolsep}{4pt}
\begin{tabular}{@{}lccccc@{}}
\toprule
\textbf{Configuration}
& \textbf{R-L}$\uparrow$
& \textbf{B@1}$\uparrow$
& \textbf{B@2}$\uparrow$
& \textbf{B@3}$\uparrow$
& \textbf{B@4}$\uparrow$ \\
\midrule

\multicolumn{6}{@{}l}{\emph{(a) Framework components}} \\
SFT baseline
& 55.03 & 51.16 & 37.99 & 29.40 & 23.46 \\
\quad + Pretraining
& 58.78 & 56.81 & 44.85 & 36.18 & 29.72 \\
\quad + Retrieval
& 60.20 & 59.39 & 46.97 & 37.95 & 31.22 \\
\quad + \textbf{RUG-RFT}
& \textbf{62.34}
& \textbf{61.12}
& \textbf{48.74}
& \textbf{39.73}
& \textbf{32.91} \\

\midrule

\multicolumn{6}{@{}l}{\emph{(b) Reward design}} \\
Retrieval-augmented SFT
& 60.20 & 59.39 & 46.97 & 37.95 & 31.22 \\
Quality Reward only
& 61.67 & 60.76 & 48.42 & 38.97 & 31.97 \\
Retrieval-Utility Reward only
& 61.37 & 60.55 & 47.93 & 38.62 & 31.66 \\
\textbf{Both rewards}
& \textbf{62.34}
& \textbf{61.12}
& \textbf{48.74}
& \textbf{39.73}
& \textbf{32.91} \\

\bottomrule
\end{tabular}
\end{table}

\subsection{Analysis}

\paragraph{Retrieval quality analysis.}\label{sec:main_analysis}

We investigate the relationship between the number of retrieved contexts, retrieval quality, and final translation performance. For each query, we vary the number of retrieved contexts $K$ and measure the token-level overlap between the retrieved texts and the ground-truth translation. The resulting scores are then averaged over all queries to reflect the overall retrieval quality. Chinese sentences are segmented using Jieba, with punctuation, empty tokens, and symbol-only tokens removed. Let $\mathcal{T}_{\mathrm{GT}}$ denote the token set of the ground-truth translation, and let $\mathcal{T}_{\mathrm{ret}}^{(K)}$ denote the token set aggregated from the top-$K$ retrieved translations. We define \emph{Retrieval Recall}, \emph{Retrieval Precision}, and \emph{Retrieval F1} as
\begin{equation}
\mathrm{Rec}_{\mathrm{ret}}(K)
=
\tfrac{
|\mathcal{T}_{\mathrm{GT}} \cap \mathcal{T}_{\mathrm{ret}}^{(K)}|
}{
|\mathcal{T}_{\mathrm{GT}}|
},~~
\mathrm{Prec}_{\mathrm{ret}}(K)
=
\tfrac{
|\mathcal{T}_{\mathrm{GT}} \cap \mathcal{T}_{\mathrm{ret}}^{(K)}|
}{
|\mathcal{T}_{\mathrm{ret}}^{(K)}|
},~~~
\mathrm{F1}_{\mathrm{ret}}(K)
=
\tfrac{
2\,\mathrm{Rec}_{\mathrm{ret}}(K)\,\mathrm{Prec}_{\mathrm{ret}}(K)
}{
\mathrm{Rec}_{\mathrm{ret}}(K)+\mathrm{Prec}_{\mathrm{ret}}(K)
}.
\end{equation}

\begin{figure*}[t]
\centering
\captionsetup[subfigure]{skip=2pt}

% =========================================================
% (a) Retrieval analysis
% =========================================================
\begin{subfigure}[t]{0.465\textwidth}
    \centering
    \includegraphics[
        height=0.62\linewidth,
        keepaspectratio
    ]{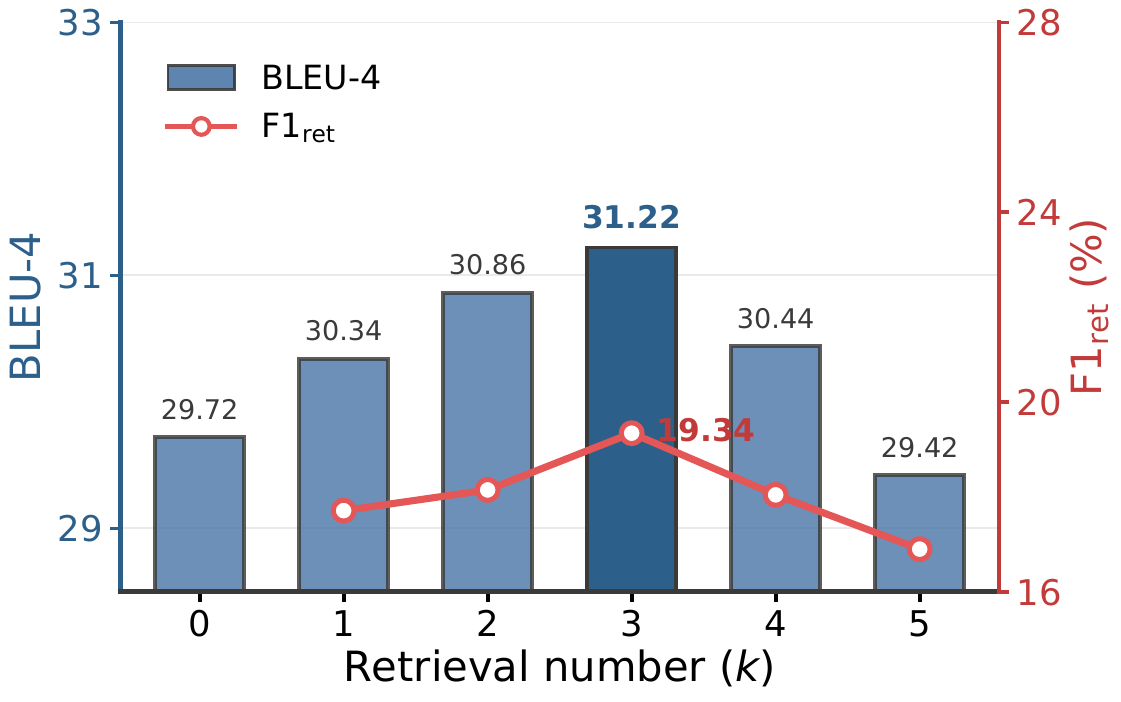}
    \caption{Retrieval quality vs.\ retrieval number}
    \label{fig:retrieval_ana}
\end{subfigure}
\hspace{0.015\textwidth}
% =========================================================
% (b) Model scaling
% =========================================================
\begin{subfigure}[t]{0.465\textwidth}
    \centering
    \includegraphics[
        height=0.62\linewidth,
        keepaspectratio
    ]{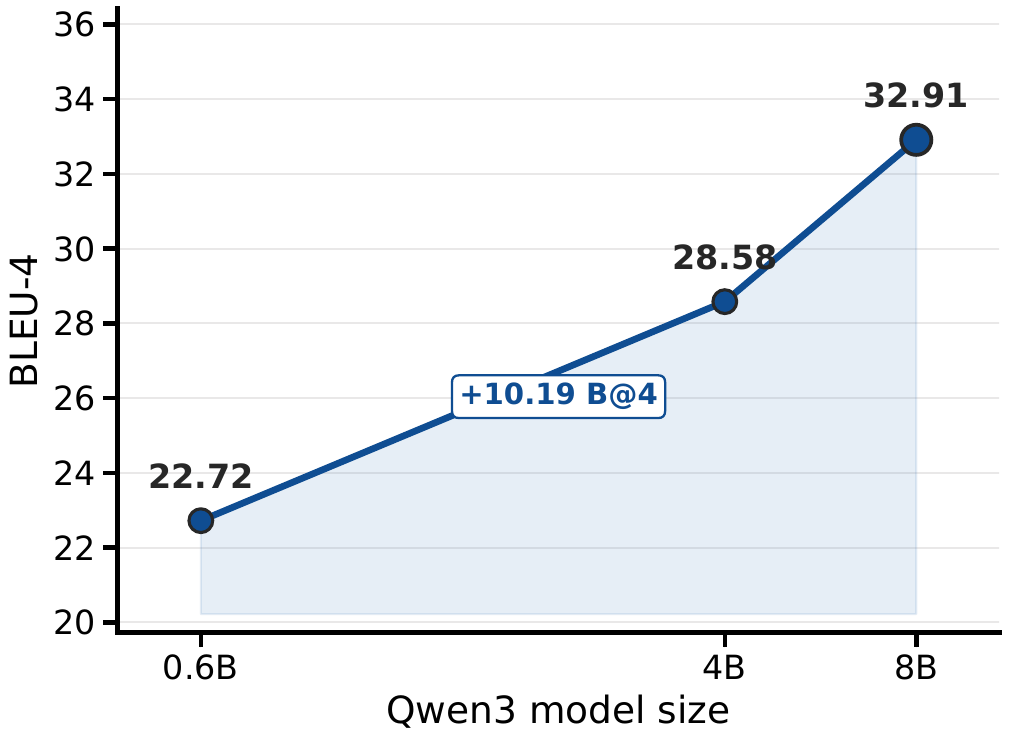}
    \caption{BLEU-4 vs.\ LLM size}
    \label{fig:model_scaling}
\end{subfigure}

\vspace{-2mm}

\caption{\small
Main analysis.
\textbf{(a)} Relationship between retrieval quality and final translation performance under different numbers of retrieved contexts.
% \textbf{(b)} BLEU-4 scaling with LLM size. Under the same training and evaluation protocol, using Qwen3 backbones of 0.6B, 4B, and 8B improves BLEU-4 from 22.72 to 28.58 and 32.91, respectively.
\textbf{(b)} BLEU-4 scaling with LLM size. We evaluate Qwen3 backbones with different parameter scales (0.6B, 4B, and 8B) under the same training and evaluation protocol.
}
\label{fig:analysis}

\vspace{-3mm}
\end{figure*}

Here, $\mathrm{Rec}_{\mathrm{ret}}$ measures how much ground-truth lexical information is covered by the retrieved contexts, while $\mathrm{Prec}_{\mathrm{ret}}$ measures how much of the retrieved lexical content is relevant to the ground truth. Their harmonic mean, $\mathrm{F1}_{\mathrm{ret}}$, captures the trade-off between retrieval coverage and retrieval noise. As shown in Fig.~\ref{fig:retrieval_ana}, as $K$ increases, $\mathrm{F1}_{\mathrm{ret}}$ first improves and then continuously decreases, and the final translation performance exhibits the same trend. This is because introducing additional less-relevant contexts can instead interfere with SLT performance. All BLEU-4 scores in Fig.~\ref{fig:retrieval_ana} are obtained from SFT models.

% ===================== Combined analysis figure =====================

\paragraph{Model scaling analysis.}

We further investigate how the scale of the language model affects translation performance. As shown in Fig.~\ref{fig:model_scaling}, under the same training and evaluation protocol, BLEU-4 increases from 22.72 to 28.58 and 32.91 when scaling the Qwen3 backbone from 0.6B to 4B and 8B, respectively. The consistent improvement shows that contemporary LLMs also exhibit stronger capabilities in sign language translation. These results further indicate that SignRAG can effectively benefit from scaling the underlying LLM.

\paragraph{Case study.}

We provide a qualitative comparison with the current SOTA model, Geo-Sign~\citep{fish2026geo}, to analyze the effect of retrieved contexts. As shown in Table~\ref{tab:qualitative-retrieval-case}, the reference sentence contains fine-grained semantic cues, including ``Traditionally'' and ``lucky money.'' Although Geo-Sign produces a partially related translation, it fails to preserve these cues. In contrast, SignRAG retrieves semantically aligned contexts that provide relevant topical and lexical cues. This further supports our finding that retrieval can improve translation performance. Additional qualitative examples are provided in Table~\ref{tab:supp_qualitative_cases} in the Appendix. Meanwhile, the bad case presented in Appendix~\ref{app:badcase} shows that noisy retrieval can mislead the model, further motivating RUG-RFT.

\begin{CJK*}{UTF8}{gbsn}
\begin{table}[ht]
\centering
\scriptsize
\setlength{\tabcolsep}{3pt}
\renewcommand{\arraystretch}{0.95}

\begin{tabular}{@{}p{0.13\linewidth}p{0.83\linewidth}@{}}
\toprule
\textbf{Source} & \textbf{Sentence} \\
\midrule

\textbf{Reference} &
红包在传统意义上也叫压岁钱。\textit{(Traditionally, a red envelope is also called ``lucky money.'')} \\

\textbf{Retrieved} &
[1] 你打算怎么花这笔压岁钱呢？\textit{(How do you plan to spend this lucky money?)}\ \ 
[2] 带着祝福的红包比攀比中的计较更有意义。\textit{(A red envelope carrying good wishes is more meaningful than one given for comparison or competition.)}\ \ 
[3] 你打算怎么花这笔压岁钱呢？\textit{(How do you plan to spend this lucky money?)} \\

\midrule

\textbf{Geo-Sign} &
红包在历史上的原意是成年礼。\textit{(Historically, the original meaning of a red envelope was a coming-of-age gift.)} \\

\textbf{SignRAG} &
红包在传统意义上是压岁钱。\textit{(Traditionally, a red envelope means ``lucky money.'')} \\

\bottomrule
\end{tabular}

\vspace{-1mm}
\caption{\small Qualitative comparison on CSL-Daily.}
\vspace{-2mm}
\label{tab:qualitative-retrieval-case}

\end{table}
\end{CJK*}

\section{Conclusion}
We propose SignRAG, a unified framework for gloss-free sign language translation. SignRAG strengthens sign representation learning through staged pretraining and leverages target-domain semantic memory to provide instance-specific linguistic cues for downstream translation. Furthermore, RUG-RFT jointly considers translation quality and retrieval utility to encourage effective use of beneficial retrieved contexts. Experiments across multiple SLT benchmarks show that SignRAG achieves new state-of-the-art performance and surpasses existing gloss-supervised methods across all reported metrics on CSL-Daily. Future work will explore more efficient and robust retrieval mechanisms, as well as knowledge transfer across datasets and languages.

% We propose SignRAG, a retrieval-augmented framework for gloss-free SLT that builds a train-split sign-to-text retrieval gallery and uses semantically related examples as auxiliary textual cues for LLM-based translation. With RUG-RFT, SignRAG regulates the model's dependence on retrieved contexts and encourages their beneficial use.
% Experiments on four SLT benchmarks show that SignRAG improves over strong gloss-free baselines under the evaluated in-domain retrieval setting.

% SignRAG still depends on the quality, coverage, and domain match of in-domain retrieval galleries.
% Noisy or superficially related contexts may mislead generation, while cross-domain, cross-signer, and cross-language retrieval remain underexplored.
% The retrieval module also adds storage and inference-time costs.
% Future work will study efficient retrieval, larger and more diverse galleries, cross-lingual transfer, human evaluation, and stronger multimodal LLMs for improving semantic grounding and context utilization in retrieval-augmented SLT.

\subsection*{AI Use Statement}
We used generative AI tools primarily to assist with writing and language polishing, including improvements to grammar, wording, clarity, and readability. We did not use generative AI tools for research design, experimental design, data analysis, or result interpretation. All AI-assisted text was carefully reviewed and revised by the authors, who take full responsibility for the final content of the manuscript.

\subsection*{Reproducibility Statement}
Although we have provided the experimental settings, training protocols, and implementation details as comprehensively as possible in the main paper and Appendix, we recognize that reproducibility is essential for impactful scientific research. To facilitate further verification and future research, upon acceptance, we will release the source code, pretrained checkpoints from different training stages, and models of different sizes. We hope these resources will support reproducible evaluation and further academic research.



\bibliography{references}
\bibliographystyle{iclr2027_conference}
\clearpage

\appendix
% 补充材料目录：显示 section 和 subsection
% 如需显示 subsubsection，将 tocdepth 改为 3
\startcontents[appendices]
\printcontents[appendices]{l}{1}{%
    \section*{APPENDIX}
    \setcounter{tocdepth}{2}%
}

\clearpage

\section{Extended Related Work}\label{app:related_work}
This section provides the full version of the related-work discussion that is condensed into a single paragraph in the main text.

\subsection{Sign Language Translation}
Sign language translation (SLT) aims to recover the semantic content conveyed by a signer from a sequence of performed signs. Unlike standard video understanding tasks, SLT must account for the unique linguistic structure of sign language. Specifically, sign sequences are more naturally aligned with gloss units, whereas spoken language sentences often follow substantially different grammatical patterns. This mismatch makes direct translation challenging. As a result, early SLT methods~\citep{SLRT, SignBT, TS-SLT, MMTLB} commonly relied on gloss supervision, typically by introducing auxiliary Sign Language Recognition modules or explicitly modeling gloss generation before sentence translation. Despite their strong performance, such methods depend heavily on costly gloss annotations, which limits scalability. To address this issue, recent work has increasingly focused on gloss-free SLT, which learns a direct mapping from sign videos to spoken language sentences. Early studies~\citep{NSLT,SLRT,TSPNet,CSGCR,GASLT} demonstrated the feasibility of this setting, although a clear performance gap from gloss-based methods remains. More recent efforts attempt to close this gap by improving visual-text representation learning. For example, GFSLT~\citep{GFSLT-VLP} adapts CLIP~\citep{CLIP} to sign language, inspiring a series of vision-language SLT models~\citep{SIGN2GPT,Fla-LLM,Sign-LLM,VAP,C2RL}. Another line of work improves SLT from the data perspective. Recent studies~\citep{YT-ASL-SLT,SSVP-SLT,Uni-Sign,zhang2024scaling,signmusketeers,jang2025lost,fish2026geo} show that pre-training on large-scale sign corpora can significantly benefit downstream translation. However, most existing efforts focus on pre-training T5-based encoder--decoder architectures, whose language priors, general knowledge, and generative capabilities remain limited compared with contemporary LLMs. Therefore, how to design effective pre-training strategies for adapting sign representations to contemporary LLM-based SLT models remains largely unexplored.

\subsection{RL for Retrieval-Enhanced Reasoning}

Retrieval-augmented generation (RAG) enhances parametric language models by
incorporating non-parametric external contexts, where retrieved evidence is
used as additional information during generation~\citep{lewis2020retrieval,
chen2022murag}. By providing explicit external references, RAG improves
knowledge utilization and reduces the dependence on purely parametric memory.
Beyond general text generation, retrieval-based approaches have also been
explored in sign language translation. However, existing retrieval-based SLT
methods~\citep{RASLT} mainly focus on gloss-based settings, where retrieval is
performed under the guidance of gloss representations. 
% Therefore, how to introduce retrieval augmentation into gloss-free SLT with modern LLM decoders
% remains underexplored.

In parallel, reinforcement learning (RL) has become an effective post-training
strategy for improving the reasoning and generation abilities of large
language models. Group Relative Policy Optimization (GRPO)
~\citep{shao2024deepseekmath,guo2025deepseek} provides an efficient RL
framework by estimating relative advantages among grouped samples without
requiring an additional value model. Recent studies have explored combining RL
with RAG to optimize retrieval-enhanced reasoning and generation.
RAG-RL~\citep{huang2025rag} improves answer generation in retrieval contexts
through reinforcement learning and curriculum strategies. R3-RAG
~\citep{li2025r3} further enables joint reasoning and retrieval optimization
with outcome and process rewards. VRAG-RL~\citep{wang2025vrag} extends this
paradigm to visual-rich scenarios by jointly optimizing retrieval, reasoning,
and generation.

These studies demonstrate the potential of RL for dynamically optimizing the
interaction between retrieval and generation. However, existing RL-based RAG
methods mainly target knowledge-intensive reasoning tasks, rather than
multimodal translation with sign language inputs. In contrast, SignRAG
investigates how large-scale sign pretraining should be organized under a
modern LLM decoder, and complements parametric adaptation with a sign-to-text
retrieval gallery and retrieval utility-guided reinforcement fine-tuning.
Unlike previous retrieval-based SLT approaches, SignRAG operates without gloss
annotations and explicitly optimizes the usefulness of retrieved translation
contexts during generation.

\section{Detailed Formulations and Derivations}
\subsection{Vision-Language Pre-training}\label{app:pretrain}
For completeness, we provide additional details of the vision-language pre-training procedure.

\subsection{Attention-weighted Skeleton--Text Similarity}
Skeleton and text features are projected into a shared $D$-dimensional space and $\ell_2$-normalized.
Given skeleton features $\{f_i\}_{i=1}^{M}$ and text features $\{t_j\}_{j=1}^{L}$, we compute token-level similarities
\begin{equation}
E_{i,j} = f_i^\top t_j.
\end{equation}

To calculate the global skeleton-to-text similarity $z_{S,T}$, a row-wise Softmax~\citep{cico} 
is applied to $\bm{E}$ to obtain the attention weights $\bm{P}$:
\begin{align}
    P_{i,j} = \frac{\exp(E_{i,j} / \tau)}{\sum_{k=1}^L \exp(E_{i,k} / \tau)},
\end{align}
where $\tau$ denotes a trainable temperature parameter. The global skeleton--text similarity is defined as
\begin{equation}
z_{S,T}=\frac{1}{M}\sum_{i=1}^{M}\sum_{j=1}^{L}P_{i,j}\,E_{i,j}.
\end{equation}
The text-to-skeleton similarity is computed analogously.

\subsection{Bidirectional InfoNCE Loss}
Given a mini-batch of $B$ paired samples, we construct batch-level similarity matrices 
and apply a bidirectional InfoNCE objective with temperature $\tau'$.

\subsection{Memory-bounded distributed retrieval.}
\label{app:retrieval_detail}

This section describes the implementation of memory-efficient retrieval in SignRAG.
Let $\mathcal{G}$ denote a training gallery containing $N$ sign sequences, each represented by frame-level features with channel dimension $C$.
Loading the entire gallery into GPU memory requires storage proportional to $\mathcal{O}(NLC)$, where $L$ denotes the padded temporal length.
Comparing a query batch against the full gallery further introduces memory costs for pairwise similarities and intermediate computations.
To support large galleries, we combine CPU-resident gallery storage, distributed sharding, and block-wise similarity computation, followed by exact distributed top-$K$ aggregation.

\paragraph{CPU-resident gallery sharding.}
With $R$ distributed workers, we partition the gallery into disjoint shards:
\begin{equation}
    \mathcal{G}
    =
    \bigcup_{r=1}^{R}\mathcal{G}^{(r)},
    \qquad
    \mathcal{G}^{(r)}
    \cap
    \mathcal{G}^{(r')}
    =
    \varnothing,
    \quad r\neq r'.
\end{equation}
Each worker is responsible for approximately $N/R$ gallery samples.
The encoded features of its local shard are retained in CPU memory and transferred to GPU only for the current similarity computation.
Thus, no worker needs to keep its entire gallery shard, or the complete gallery, in GPU memory.

\paragraph{Block-wise similarity computation.}
We divide the queries into chunks $\mathcal{Q}_{b}$ of at most $B_q$ sequences and each local gallery shard into chunks $\mathcal{G}^{(r)}_{c}$ of at most $B_g$ sequences:
\begin{equation}
    |\mathcal{Q}_{b}|\leq B_q,
    \qquad
    |\mathcal{G}^{(r)}_{c}|\leq B_g.
\end{equation}
Each query chunk is broadcast to all workers.
Worker $r$ then compares it against the chunks of its local gallery shard in turn.
For the current query--gallery block, the aggregated similarities are
\begin{equation}
    A^{(r)}_{b,c}[u,v]
    =
    S\left(Q^{b}_{u},G^{(r,c)}_{v}\right),
    \qquad
    A^{(r)}_{b,c}
    \in
    \mathbb{R}^{|\mathcal{Q}_{b}|\times|\mathcal{G}^{(r)}_{c}|},
\end{equation}
where $S(\cdot,\cdot)$ is the attention-weighted sign--sign similarity defined in Eq.~\ref{eq:sign_sign_sim}.
After evaluating a block, its similarity scores are transferred to CPU memory, and the GPU storage occupied by the current gallery chunk can be released before loading the next one.
The query chunk remains available while the worker processes its local gallery.
Scores from successive gallery chunks are collected on CPU for local candidate selection.

\paragraph{GPU memory usage.}
Let $L$ bound the padded temporal lengths of the active query and gallery chunks.
The resident retrieval tensors on each GPU require
\begin{equation}
\label{eq:retrieval_memory}
    \mathcal{M}_{\mathrm{resident}}
    =
    \mathcal{O}\left(
        (B_q+B_g)LC+B_qB_g
    \right).
\end{equation}
The first term accounts for the active query and gallery features, and the second accounts for their aggregated similarity scores.
This bound excludes model parameters, temporary intermediate activations, and kernel workspace; the actual peak GPU memory also depends on the implementation of $S(\cdot,\cdot)$.
For fixed chunk sizes and sequence lengths, the resident GPU storage is independent of the total gallery size $N$.
The chunk sizes can therefore be adjusted to the available device memory.
CPU storage still grows with the local gallery size and the similarity scores retained for candidate selection, and retrieval still evaluates every eligible query--gallery pair.

\paragraph{Candidate filtering.}
During training-set retrieval, we exclude the query itself, identified by its sample identity, and gallery samples whose reference translations are identical to that of the query.
These exclusions prevent trivial retrieval of the query or entries with identical supervision.
Let $y_x$ denote the reference translation of sample $x$.
The eligible local gallery for a training query $q$ is
\begin{equation}
    \mathcal{E}^{(r)}(q)
    =
    \left\{
        g\in\mathcal{G}^{(r)}
        \;\middle|\;
        g\neq q,\; y_g\neq y_q
    \right\}.
\end{equation}
For development and test queries, retrieval is performed against the training gallery without using the query reference translation, so
$\mathcal{E}^{(r)}(q)=\mathcal{G}^{(r)}$.
All exclusions are applied before local top-$K$ selection to ensure that eligible candidates are not displaced by entries that would subsequently be removed.

\paragraph{Distributed top-$K$ aggregation.}
After evaluating its local gallery, worker $r$ selects
\begin{equation}
    \mathcal{C}^{(r)}_{K}(q)
    =
    \operatorname{TopK}_{g\in\mathcal{E}^{(r)}(q)}
    S(q,g),
\end{equation}
returning all eligible entries if fewer than $K$ are available.
Only the selected sample identifiers and their similarity scores are communicated for final aggregation.
The global retrieved set is
\begin{equation}
\label{eq:distributed_topk}
    \mathcal{R}_{K}(q)
    =
    \operatorname{TopK}_{
        g\in\bigcup_{r=1}^{R}\mathcal{C}^{(r)}_{K}(q)
    }
    S(q,g).
\end{equation}
This aggregation preserves exact exhaustive retrieval.
Under a consistent ordering for tied scores, any candidate outside its shard's local top-$K$ is preceded by at least $K$ eligible candidates from that shard and therefore cannot belong to the global top-$K$.
Consequently, selecting from the union of local candidates yields the same result as selecting directly from the full eligible gallery.
Final aggregation requires at most $RK$ candidates per query, without gathering the gallery features or the full similarity matrix across workers.

\paragraph{Communication and CPU fallback.}
In our implementation, variable-length query payloads and retrieved candidate metadata are communicated through a CPU-side Gloo process group.
This avoids the additional GPU-memory allocation associated with object collectives over NCCL.
If a query--gallery block still exceeds the available GPU memory, the implementation supports a CPU fallback that evaluates the same similarity function $S(\cdot,\cdot)$.
The fallback changes the execution device while preserving the scoring definition, eligibility rules, and exhaustive retrieval procedure.

\section{More Implementation Details}\label{sec:More Details}
To enable a thorough understanding of SignRAG, this section presents more implementation details.

\subsection{Computing Infrastructure}
The entire experimental process was conducted on a computational infrastructure comprising an AMD EPYC 7502 32-Core Processor, 4 NVIDIA RTX 4090D GPUs with 48GB of VRAM, running the Ubuntu 20.04 operating system, with Python 3.10, CUDA 11.8, transformers 4.51.3 and PyTorch 2.1.2.

\subsection{Keypoint Extraction.}
Following the preprocessing protocol adopted in CoSign~\citep{cosign}, we extract frame-level human keypoints from sign-language videos using the off-the-shelf MMPose estimator. To focus on motion and appearance cues that are most relevant to sign understanding, we retain a compact set of keypoints covering the upper body, hands, mouth, and face. Specifically, the selected keypoints include 9 body joints, 21 joints for each hand, 8 mouth landmarks, and 18 facial landmarks, resulting in 77 keypoints per frame. Additional implementation details follow CoSign~\citep{cosign}.

% Required packages:
% \usepackage{booktabs}
% \usepackage{tabularx}
% \usepackage{array}

\begin{table*}[t]
\centering
\caption{
Implementation details of the four training stages.
LR denotes learning rate, FFN denotes feed-forward network, and RFT denotes reinforcement fine-tuning.
}
\label{tab:training_settings_supp}

\footnotesize
\setlength{\tabcolsep}{4.5pt}
\renewcommand{\arraystretch}{1.12}

% Left-aligned columns with automatic line wrapping
\newcolumntype{L}[1]{>{\raggedright\arraybackslash}p{#1}}
\newcolumntype{Y}{>{\raggedright\arraybackslash}X}

\begin{tabularx}{\textwidth}{
@{}
L{0.145\textwidth}
L{0.195\textwidth}
L{0.295\textwidth}
Y
@{}
}
\toprule
\textbf{Stage}
& \textbf{Optimization}
& \textbf{Model Configuration}
& \textbf{Data / Training Details} \\
\midrule

\textbf{Pretraining I}

\emph{Sign--Text Pretraining}
&
\textbf{Epochs:} 200

\textbf{Optimizer:} SGD

\textbf{LR:} 0.01

\textbf{Weight decay:} 0.001

\textbf{Scheduler:} cosine

\textbf{Warmup:} 20 epochs

\textbf{Batch size:} 8
&
\textbf{Text encoder/decoder:} first three layers of mBART

\textbf{Skeleton encoder:} three-layer Transformer

\textbf{Attention heads:} 8

\textbf{Hidden size:} 512

\textbf{FFN dimension:} 2048

\textbf{Contrastive dimension:} 1024
&
\textbf{Training corpus:} CSL-News

\textbf{Vocabulary size:} 11,540
\\

\midrule

\textbf{Pretraining II}

\emph{Large-Scale Sign MLLM Pretraining}
&
\textbf{Epochs:} 50

\textbf{Optimizer:} AdamW

\textbf{LR:} $1 \times 10^{-4}$

\textbf{Batch size:} 8

\textbf{Objective:} cross-entropy
&
\textbf{Base LLM:} Qwen3-8B

\textbf{Projector:} MLP with GELU

\textbf{Projection:} $512 \rightarrow 4096$

\textbf{LoRA targets:}
\mbox{\texttt{q\_proj}},
\mbox{\texttt{v\_proj}}

\textbf{LoRA rank:} 16

\textbf{Scaling factor:} 32

\textbf{Dropout:} 0.3
&
\textbf{Training corpora:} CSL-News / YouTube-ASL

The sign encoder, multimodal projector, and LLM are jointly pretrained.
\\

\midrule

\textbf{SFT}

\emph{Retrieval-Augmented SFT}
&
\textbf{Epochs:} 10

\textbf{Optimizer:} AdamW

\textbf{LR:} $1 \times 10^{-4}$

\textbf{Batch size:} 4

\textbf{Objective:} cross-entropy
&
Stage-II LoRA weights are merged into the base LLM, followed by a new set of LoRA modules.

\textbf{LoRA targets:}
\mbox{\texttt{q\_proj}},
\mbox{\texttt{v\_proj}}

\textbf{LoRA rank:} 16

\textbf{Scaling factor:} 32

\textbf{Dropout:} 0.3
&
\textbf{Training data:} downstream SLT datasets

Retrieved texts and sign features are jointly used as multimodal input.

\textbf{Prompt:} Fig.~\ref{fig:prompt}
\\

\midrule

\textbf{RFT}

\emph{Retrieval Utility-Guided RFT}
&
\textbf{Epochs:} 2

\textbf{Optimizer:} AdamW

\textbf{LR:} $1 \times 10^{-5}$

\textbf{Batch size:} 8

\textbf{Responses per input:} 8
&
SFT LoRA weights are merged with the pretrained LLM weights, followed by a new set of LoRA modules.

\textbf{LoRA targets:}
\mbox{\texttt{q\_proj}},
\mbox{\texttt{v\_proj}}

\textbf{LoRA rank:} 64

\textbf{Scaling factor:} 128

\textbf{Dropout:} 0.05
&
\textbf{Optimization:} GRPO

\textbf{Rewards:} sentence-level quality reward and retrieval-utility reward

The same prompt template as SFT is used.

% Sentence-final punctuation is appended to very short sentences before BLEU-4-based reward computation.
\\

\bottomrule
\end{tabularx}
\end{table*}

\subsection{Pretraining(Stage~I): Sign-text pre-training.}
In the first stage, we train a compact sign-text model to align skeleton-based sign representations with textual semantics. Following GFSLT-VLP~\citep{GFSLT-VLP}, the text encoder and decoder are initialized by reusing the first three layers of mBART, while the skeleton encoder adopts the same three-layer Transformer structure. Each Transformer block uses 8 attention heads, a hidden size of 512, and a feed-forward dimension of 2048. The model is optimized for 200 epochs using SGD with a learning rate of 0.01 and a weight decay of 0.001. A cosine learning rate schedule is applied with 20 warmup epochs. The batch size is set to 8. For CSL-News, we follow prior work and reduce the vocabulary to a task-specific subset of 11,540 tokens. The contrastive feature dimension for both visual and textual representations is set to 1024.

\subsection{Pretraining(Stage~II): Large-scale sign MLLM pretraining.}
We use Qwen3-8B as the base language model. Following the design of LLaVA~\citep{llava}, an MLP projector with GELU activation maps the 512-dimensional visual features into the 4096-dimensional embedding space of the LLM. To enable parameter-efficient adaptation, we apply LoRA to the query and value projection matrices in the self-attention layers of the LLM. The LoRA rank, scaling parameter, and dropout rate are set to 16, 32, and 0.3, respectively. The model is trained for 50 epochs using the AdamW optimizer with a learning rate of $1 \times 10^{-4}$ and a batch size of 8. Cross-entropy loss is used as the training objective.

\subsection{SFT: Retrieval-Augmented Supervised Fine-Tuning.}
\begin{figure}[ht]
\centering
\includegraphics[width=1\linewidth]{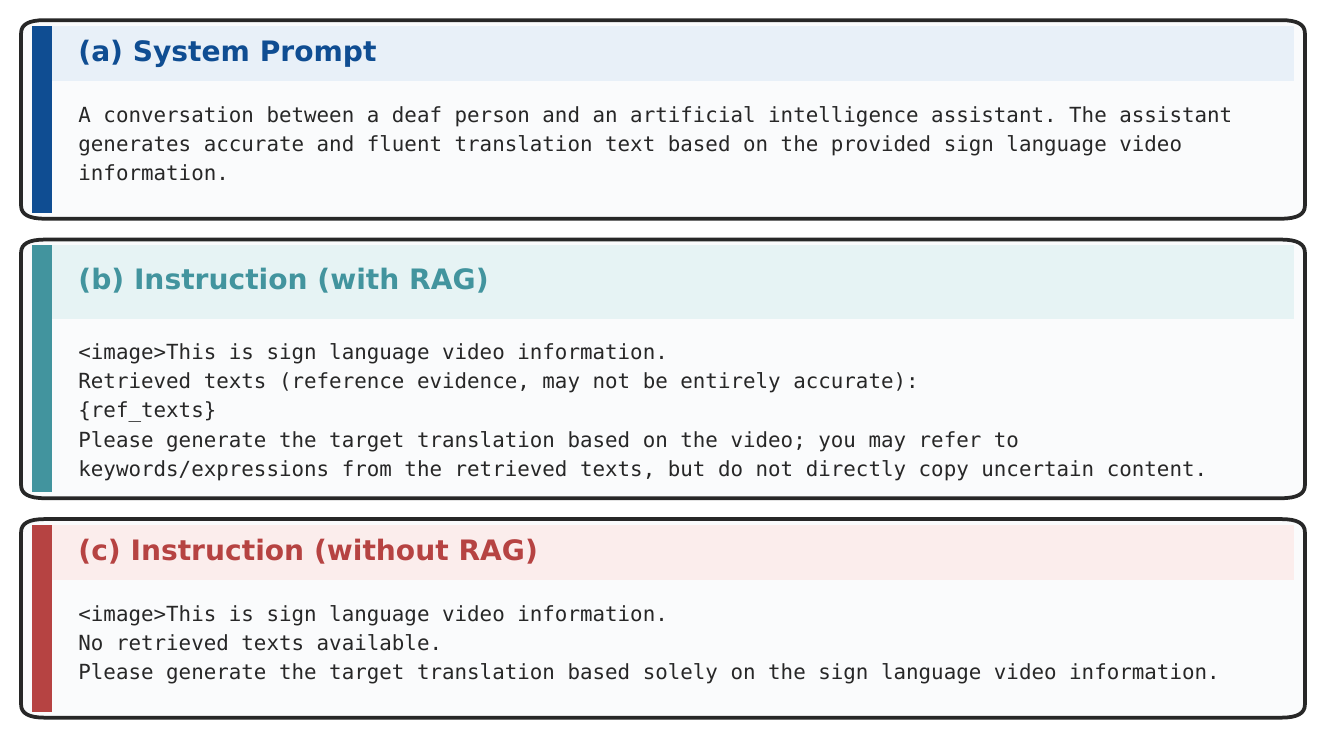}
\vspace{-3mm}
\caption{
\small 
System prompt and instruction template. 
Here, \textit{with RAG} denotes the setting where retrieved contexts are provided to the LLM, 
while \textit{without RAG} denotes the setting where no retrieved contexts are provided.
}
\label{fig:prompt}
\vspace{-3mm}
\end{figure}

After large-scale sign MLLM pretraining, we merge the learned LoRA parameters into the base LLM to obtain the final pretrained LLM checkpoint. We then attach a new set of LoRA modules using the same LoRA configuration, including the rank, scaling factor, and dropout rate. The main difference in this stage is that the training data are switched from large-scale pretraining corpora, such as CSL-News and YouTube-ASL, to downstream SLT datasets. Retrieved texts are additionally incorporated and combined with the sign features to construct the multimodal input. The system prompt and instruction template used in this stage are shown in Fig.~\ref{fig:prompt}. During SFT, the model is trained for 10 epochs using AdamW with a learning rate of $1 \times 10^{-4}$. The batch size is set to 4, and cross-entropy loss is used as the training objective.

\subsection{RFT: Retrieval Utility-Guided Reinforcement Fine-Tuning.}
After supervised fine-tuning, we merge the learned LoRA parameters with our pretrained LLM weights to obtain the SignRAG-SFT checkpoint. We then attach a new set of LoRA modules for reinforcement fine-tuning. In this stage, the LoRA rank is increased to 64, the scaling factor is set to 128, and the dropout rate is reduced to 0.05. The model is trained for 2 epochs using AdamW with a learning rate of $1 \times 10^{-5}$ and a batch size of 8. Following GRPO, we sample 8 candidate responses for each input and compute group-relative advantages based on both the sentence-level quality reward and the retrieval-utility reward. The same prompt template used in the SFT stage is adopted for RFT. For datasets containing very short sentences, we append sentence-final punctuation before reward computation to reduce the instability of BLEU-4-based training signals.

\subsection{Identical-text filtering in training retrieval.}\label{app:identical_retrieval}
\algrenewcommand\algorithmicrequire{\textbf{Input:}}
\algrenewcommand\algorithmicensure{\textbf{Output:}}

We use CSL-Daily as an example to illustrate the retrieval procedure.
The retrieval gallery $\mathcal{G}$ consists of all training samples, where each gallery entry is represented as
$\mathcal{G}_i=(g_i,t_i,id_i)$, containing the sign sample, its translation, and sample ID.
Given a query $q$, all gallery samples are first ranked according to their visual similarity, followed by sample-level and text-level filtering.
The complete procedure is summarized in Algorithm~\ref{alg:retrieval_context}.

\begin{algorithm}[t]
\caption{Retrieval Context Construction}
\label{alg:retrieval_context}
\small
\begin{algorithmic}[1]

\Require Query $q$ with $id_q$ and $t_q$;
gallery $\mathcal{G}=\{(g_i,t_i,id_i)\}_{i=1}^{N}$;
similarity function $S(\cdot,\cdot)$; retrieval number $K$

\Ensure Retrieved contexts $\mathcal{R}_{K}(q)$

\State Compute visual similarities:
\Statex \hspace{\algorithmicindent}
$
s_i = S(q,g_i), \qquad i=1,\ldots,N
$

\State Rank gallery samples in descending order:
\Statex \hspace{\algorithmicindent}
$
\pi = \operatorname{Argsort}_{i}(s_i)
$

\State Initialize $\mathcal{R}_{K}(q) \gets \varnothing$

\For{$i \in \pi$}

    \If{$id_i = id_q$}
        \State \textbf{continue}
    \EndIf

    \If{\textsc{TrainRetrieval} $\land$ $t_i=t_q$}
        \State \textbf{continue}
    \EndIf

    \State
    $\mathcal{R}_{K}(q)
    \gets
    \mathcal{R}_{K}(q)\cup\{t_i\}$

    \If{$|\mathcal{R}_{K}(q)| = K$}
        \State \textbf{break}
    \EndIf

\EndFor

\State \Return $\mathcal{R}_{K}(q)$

\end{algorithmic}
\end{algorithm}

During training-set self-retrieval, identical-text filtering is introduced to prevent the target sentence of the query from being directly provided as retrieval context. Without this filtering, samples sharing the same target annotation may dominate the retrieved candidates, allowing the translation model to exploit a trivial shortcut by copying the retrieved target sentence. More importantly, this would introduce a mismatch between training and dev/test inference, since the exact target translation of a query is generally unavailable in the retrieval gallery at inference time.The effectiveness of identical-text filtering on CSL-Daily is evaluated in Table~\ref{tab:identical_text_filtering}. Only 2 test samples share identical translations with the training set, suggesting that the improvements of SignRAG are not driven by a trivial retrieval shortcut that directly matches test queries to training samples with the same target translation.

\begin{table}[ht]
\centering
\caption{Effect of identical-text filtering during training-set self-retrieval on CSL-Daily.}
\label{tab:identical_text_filtering}
\begin{tabular}{lc}
\toprule
Setting & BLEU-4 \\
\midrule
SignRAG-SFT w/o identical-text filtering & 11.21 \\
SignRAG-SFT w/ identical-text filtering  & 31.22 \\
\bottomrule
\end{tabular}
\end{table}

\section{Further Analyses} \label{sec:Further Analyses}

\subsection{Analysis of different pretraining strategies.}
We compare three pretraining strategies. \emph{Encoder-only pretraining} initializes only the sign encoder from Stage I before downstream SFT. \emph{End-to-end pretraining} directly pretrains a randomly initialized sign encoder jointly with the LLM. \emph{Hierarchical pretraining} first learns a language-aware sign encoder and then jointly pretrains the sign encoder, multimodal projector, and LLM.
As shown in Table~\ref{tab:pretraining_strategy}, hierarchical pretraining substantially outperforms direct end-to-end pretraining on CSL-Daily, achieving BLEU-4 and ROUGE-L scores of 29.72 and 58.78, compared with 9.53 and 38.70, respectively. These results highlight the importance of learning semantically grounded sign representations before integrating them with an LLM.

\begin{table}[ht]
\centering
\caption{Comparison of pretraining strategies. Results on CSL-News and CSL-Daily correspond to pretraining and downstream SFT evaluation, respectively.}
\label{tab:pretraining_strategy}
\setlength{\tabcolsep}{4.5pt}
\begin{tabular}{lccccc}
\toprule
\textbf{Pretraining Strategy}
& \textbf{R-L$\uparrow$}
& \textbf{B@1$\uparrow$}
& \textbf{B@2$\uparrow$}
& \textbf{B@3$\uparrow$}
& \textbf{B@4$\uparrow$} \\
\midrule
\multicolumn{6}{l}{\emph{Pretraining evaluation on CSL-News}} \\
Sign Encoder + Shallow Transformer
& 36.31 & 41.58 & 30.24 & 23.51 & 19.02 \\
\midrule
\multicolumn{6}{l}{\emph{Downstream SFT evaluation on CSL-Daily}} \\
Encoder-only Pretraining
& 22.51 & 18.22 & 8.76 & 5.02 & 3.20 \\
End-to-End Pretraining
& 38.70 & 32.74 & 19.87 & 13.30 & 9.53 \\
\textbf{Hierarchical Pretraining (Ours)}
& \textbf{58.78} & \textbf{56.81} & \textbf{44.85} & \textbf{36.18} & \textbf{29.72} \\
\bottomrule
\end{tabular}
\end{table}

\subsection{Directly Fine-Tuning a Video LLM.}
% 审稿人都会问的问题
Constructing large-scale sign language datasets remains challenging due to the high cost of data collection and annotation, as well as the limited availability of annotators with professional sign language expertise. Consequently, a substantial cross-modal semantic gap persists between sign videos and natural language text. Video LLMs have demonstrated strong capabilities in temporal modeling and natural language generation. Adapting such models with a limited number of sign--text pairs therefore provides a potentially effective way to leverage their existing video understanding and language modeling capabilities for cross-modal alignment, while reducing the dependence on large-scale sign language annotations and costly training. From this perspective, adapting general-purpose Video LLMs to sign language translation represents a promising research direction.

\begin{figure}[ht]
    \centering
    \includegraphics[width=0.4\linewidth]{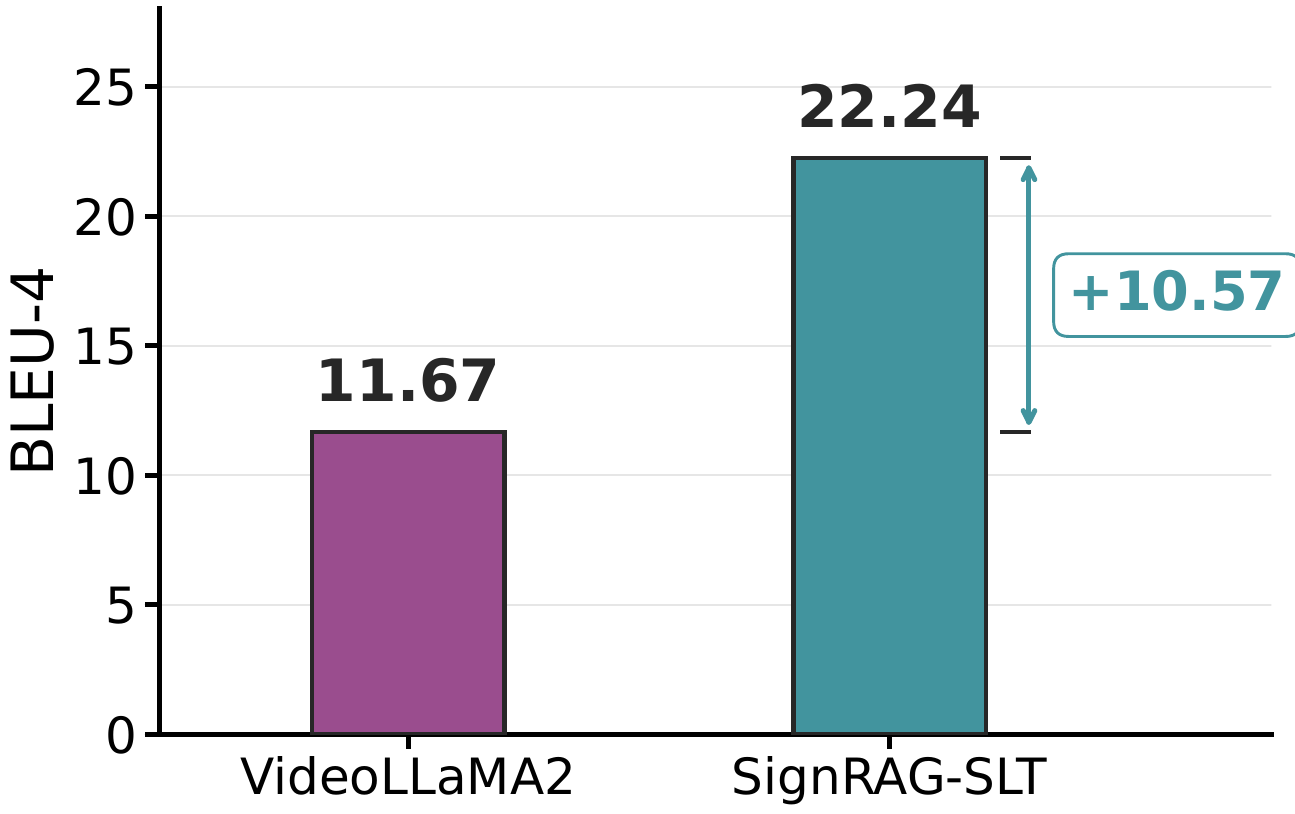}
    \vspace{-2mm}
    \caption{Performance comparison between directly fine-tuning VideoLLaMA2 and the SignRAG-SFT model.}
    \label{fig:VideoLLaMA2}
    \vspace{-3mm}
\end{figure}

To ensure a fair comparison and isolate the effect of the model architecture as much as possible, we strictly follow the official training configuration of VideoLLaMA2, keeping all training settings and hyperparameters unchanged while replacing only the original video data with sign language data. As shown in Fig.~\ref{fig:VideoLLaMA2}, our preliminary experiments reveal a substantial performance gap between directly fine-tuning VideoLLaMA2 (7B)~\citep{videollama2} on sign--text pairs and our SFT model. We attribute this gap in part to the distinctive nature of sign language, whose semantics are substantially more structured and abstract than those of general video content. The visual encoder of a general-purpose Video LLM is not specifically designed to model the fine-grained visual patterns underlying sign language and therefore struggles to learn sufficiently discriminative and linguistically relevant sign representations. Moreover, compared with generic video understanding, SLT requires visual semantics at a much finer granularity, demanding accurate modeling of handshape, motion, spatial location, non-manual cues, and their temporal compositions. Capturing such subtle linguistic information remains challenging for general-purpose video encoders.

These observations also motivate a central goal of our work: to move beyond the direct adaptation of existing general-purpose Video LLMs and toward a multimodal foundation model with stronger sign representation learning and language modeling capabilities for SLT. Although the current analysis is based on a relatively limited experimental setting, further exploring Video LLMs with stronger visual backbones, larger model scales, and more extensive sign-language-oriented pre-training remains an important direction for future work.

\subsection{Why Does Hierarchical Pretraining Benefit Sign MLLMs?}
\label{sec:supp_hierarchical_pretraining}

We further analyze why hierarchical pretraining is particularly important when replacing conventional pretrained language models with stronger LLMs. Empirically, we observe an interesting phenomenon: directly coupling a sign encoder with a powerful LLM can perform substantially worse than using a T5-based architecture, whereas pretraining the sign encoder before multimodal joint training significantly improves performance.

We attribute this phenomenon to \textbf{Cross-Modal Optimization Imbalance}~\citep{wang2020makes,huang2022modality,peng2022balanced}. At the beginning of end-to-end multimodal training, the sign encoder is either randomly initialized or only weakly aligned with linguistic semantics, while the LLM already possesses strong language modeling capability and mature linguistic priors. Therefore, the sign representation
\begin{equation}
Z_s = E_{\mathrm{sign}}(X)
\end{equation}
may still mainly encode low-level motion patterns, whereas the LLM can already partially fit the target sentence through its pretrained language knowledge. This imbalance can lead to
\begin{equation}
\underbrace{\text{Weak Sign Features}}_{\text{visual pathway}}
+
\underbrace{\text{Strong Language Prior}}_{\text{LLM}}
\rightarrow
\text{Language-Dominated Optimization}.
\end{equation}
As a result, the model may reduce the generation loss without sufficiently improving the semantic quality of the sign representation.

There are two main reasons for this behavior. First, the sign encoder only receives indirect supervision from the final generation objective:
\begin{equation}
\mathcal{L}_{\mathrm{gen}}
\rightarrow
\mathrm{LLM}
\rightarrow
\mathrm{Projector}
\rightarrow
\mathrm{Sign~Encoder}.
\end{equation}
Hence, the gradient signal reaching the sign encoder is less direct than explicit sign--text semantic supervision. Second, the multimodal projector may partially compensate for immature sign representations by mapping them into the LLM latent space, rather than forcing the sign encoder itself to develop linguistically meaningful features. Consequently, end-to-end optimization may successfully reduce the generation loss while still producing weak visual grounding. In other words,
\begin{equation}
\text{Lower Generation Loss}
\not\Rightarrow
\text{Better Sign Representation}.
\end{equation}

This issue is less pronounced for T5-based SLT models. Compared with modern LLMs, T5-style language models generally provide weaker linguistic priors and therefore cannot rely as heavily on language-side shortcuts. To minimize the translation objective, they must depend more strongly on visual evidence from the sign encoder, resulting in tighter visual-to-language coupling. This explains why a T5-based model can remain relatively stable even when the sign encoder and language model are trained jointly.

In contrast, stronger LLMs require more semantically mature modality representations before multimodal optimization. Our hierarchical pretraining explicitly addresses this requirement. In the first stage, we perform sign--text alignment:
\begin{equation}
E_{\mathrm{sign}}
\xrightarrow{\text{Sign--Text Alignment}}
Z_s^{\mathrm{sem}},
\end{equation}
where $Z_s^{\mathrm{sem}}$ is encouraged to encode linguistically structured sign semantics rather than only motion patterns. In the second stage, the aligned sign encoder is integrated with the multimodal projector and LLM:
\begin{equation}
Z_s^{\mathrm{sem}}
\rightarrow
\mathrm{Projector}
\rightarrow
\mathrm{LLM}.
\end{equation}
This decomposition transforms a difficult joint optimization problem into two progressively learned objectives:
\begin{equation}
\boxed{
\text{Sign Representation Learning}
\rightarrow
\text{Cross-Modal Generation}
}
\end{equation}
The first stage focuses on \emph{what the sign means}, while the second stage focuses on \emph{how to express the acquired sign semantics in natural language}.

Overall, these results suggest that the advantage of a stronger LLM can only be fully realized when it is paired with a sufficiently strong modality interface. More generally, effective Sign MLLM pretraining depends not only on language-model capacity, but also on the quality of modality representations, cross-modal alignment, and optimization order:
\begin{equation}
\boxed{
\text{MLLM Performance}
\propto
\text{Representation Quality}
+
\text{Alignment Quality}
+
\text{Optimization Strategy}.
}
\end{equation}
This observation motivates our hierarchical design and explains why directly scaling the language backbone does not necessarily lead to better sign language translation.

\subsection{Dataset statistics.}\label{app:dataset_statistics}

\begin{table}[ht]
\centering
\caption{Statistics of the pretraining and fine-tuning datasets. For YouTube-ASL, we report both the original dataset size and the number of samples available to us.}
\label{tab:dataset_statistics}
\begin{tabular}{@{}lllr@{}}
\toprule
Stage & Dataset & Sign language & \# Samples \\
\midrule
\multirow{3}{*}{Pretraining}
& CSL-News \citep{Uni-Sign} & Chinese & 722,715 \\
& YouTube-ASL \citep{YT-ASL-SLT} (original) & American & 530,161 \\
& YouTube-ASL (available) & American & 434,953 \\
\midrule
\multirow{4}{*}{Fine-tuning}
& How2Sign \citep{how2sign} & American & 35,172 \\
& OpenASL \citep{OpenASL} & American & 98,417 \\
& CSL-Daily \citep{SignBT} & Chinese & 20,654 \\
& PHOENIX-2014T\citep{NSLT} & German & 8,257 \\
\bottomrule
\end{tabular}
\end{table}

Table~\ref{tab:dataset_statistics} summarizes the datasets used for pretraining and fine-tuning, all of which consist of sentence-level sign--text pairs. For YouTube-ASL, we successfully obtained 434,953 of the 530,161 reported samples, corresponding to 82.04\% of the original dataset. Some of the provided YouTube links have become invalid over time, preventing access to the corresponding source videos. We therefore use the available subset for pretraining.

\subsection{Metric overfitting analysis.}\label{sec:more_metrics}
\begin{wraptable}{r}{0.4\linewidth}
\vspace{-4mm}
\centering
\caption{\small
Additional automatic metrics on CSL-Daily for checking metric overfitting.
SFT and RFT denote SignRAG-SFT and SignRAG-RFT; R-L, B@4, and BSim denote ROUGE-L, BLEU-4, and BERTSim.
}
\label{tab:additional_metrics}
\vspace{-2mm}
\setlength{\tabcolsep}{3pt}
\renewcommand{\arraystretch}{1.0}
\small
\begin{tabular}{lcccc}
\toprule
 & \textbf{R-L$\uparrow$} & \textbf{B@4$\uparrow$} & \textbf{chrF$\uparrow$} & \textbf{BSim$\uparrow$} \\
\midrule
SFT & 60.20 & 31.22 & 26.93 & 82.10 \\
RFT & \textbf{62.34} & \textbf{32.91} & \textbf{28.42} & \textbf{82.96} \\

\bottomrule
\end{tabular}
\vspace{-7mm}
\end{wraptable}
Since RUG-RFT optimizes sentence-level BLEU and ROUGE rewards, we further evaluate SignRAG with chrF~\citep{popovic2015chrf} and BERTSim, a BERT-based semantic similarity score computed with \textit{bert-base-chinese}.\footnote{https://huggingface.co/google-bert/bert-base-chinese}
As shown in Table~\ref{tab:additional_metrics}, SignRAG-RFT improves over SignRAG-SFT on ROUGE-L, BLEU-4, chrF, and BERTSim.
These complementary metrics suggest that the gains are not limited to direct BLEU/ROUGE-oriented optimization.

\subsection{Transferability across Chinese SLT benchmarks.}
To further evaluate whether our pretraining strategy generalizes beyond a specific Chinese SLT dataset, we transfer the pretrained model to CE-CSL~\citep{zhu2024chinese}, a continuous sign language benchmark collected under complex environments. As shown in Table~\ref{tab:ce_csl}, our \texttt{Pretrain$\rightarrow$SFT} consistently outperforms Uni-Sign and Geo-Sign across all metrics. In particular, it achieves 24.44 BLEU-4 and 54.55 ROUGE-L, surpassing the strongest baselines by 2.37 and 2.54 points, respectively. These results demonstrate that the learned representations can effectively transfer across different Chinese SLT benchmarks, supporting the generalizability of our pretraining strategy.

\begin{table}[ht]
\centering
\caption{
\textbf{Comparison on the CE-CSL benchmark.}
We further evaluate the adaptability of different pretrained SLT models on CE-CSL.
Uni-Sign and Geo-Sign are included as representative baselines with publicly available pretrained weights.
The results demonstrate the capability of our \texttt{Pretrain$\rightarrow$SFT} to transfer and adapt to different downstream SLT benchmarks.
}
\label{tab:ce_csl}
\setlength{\tabcolsep}{7pt}
\begin{tabular}{lccccc}
\toprule
\textbf{Method}
& \textbf{B@1$\uparrow$}
& \textbf{B@2$\uparrow$}
& \textbf{B@3$\uparrow$}
& \textbf{B@4$\uparrow$}
& \textbf{R-L$\uparrow$} \\
\midrule
Uni-Sign
& 51.23 & 37.14 & 27.89 & 21.66 & 52.01 \\

Geo-Sign
& 51.81 & 37.22 & 28.31 & 22.07 & 50.24 \\

\midrule
\textbf{\texttt{Pretrain$\rightarrow$SFT} (Ours)}
& \textbf{53.18} & \textbf{39.14} & \textbf{30.27} & \textbf{24.44} & \textbf{54.55} \\
\bottomrule
\end{tabular}
\end{table}

\subsection{Failure Case (Harmful Reliance on Retrieved Contexts).}\label{app:badcase}

Table~\ref{tab:retrieval_overreliance_case} presents a representative failure case illustrating harmful reliance on retrieved contexts. Although the \texttt{Pretrain$\rightarrow$SFT} variant, which is fine-tuned using only visual input without retrieval, correctly translates the input as \textit{The weather is nice today}, SignRAG-SFT is misled by the incorrect retrieved context and generates \textit{The supermarket is holding an event}. After RFT, SignRAG-RFT recovers the correct translation, indicating that retrieval utility-guided reinforcement fine-tuning can reduce harmful reliance on misleading retrieved information.

\begin{table}[ht]
\centering
\caption{Qualitative example illustrating harmful reliance on retrieved contexts and the effect of RFT. Red text highlights misleading or incorrectly generated content.}
\label{tab:retrieval_overreliance_case}

\small
\setlength{\tabcolsep}{5pt}
\renewcommand{\arraystretch}{1.12}

\begin{tabular}{@{}ll@{}}
\toprule
\textbf{Source} & \textbf{Text} \\
\midrule

Ground-Truth
& The weather is nice today. \\

Retrieved Text
& The \textcolor{red}{supermarket} is \textcolor{red}{closed} today. \\

\addlinespace[1pt]

\texttt{Pretrain$\rightarrow$SFT}
& The weather is nice today. \\

SignRAG-SFT
& The \textcolor{red}{supermarket} is \textcolor{red}{holding an event}. \\

SignRAG-RFT
& The weather is nice today. \\

\bottomrule
\end{tabular}

\end{table}

\section{Efficiency, Computational Cost, and Retrieval Scalability}
\label{sec:efficiency}

In this section, we provide a systematic analysis of the computational overhead, retrieval scalability, memory consumption, and training cost of SignRAG.
Unless otherwise specified, all efficiency experiments are conducted using Qwen3-8B on four NVIDIA RTX 4090D GPUs.
For a fair comparison, the Non-RAG baseline (\texttt{Pretrain$\rightarrow$SFT}) and SignRAG-SFT are evaluated using the same hardware, batch size, numerical precision, and decoding strategy.
Since inference latency reported in prior work is often measured under different hardware and implementation settings, we focus on the incremental cost of SignRAG relative to this hardware-controlled baseline.

\subsection{Online SLT Inference Efficiency}
\label{sec:online_efficiency}

A key property of SignRAG is that retrieval is performed \emph{offline}.
Specifically, gallery search is conducted once before training or evaluation, and the retrieved results are stored in the data manifests.
Therefore, no full-gallery search is performed during the deployed \texttt{model.generate} path.
The additional online inference cost mainly arises from the increased input length introduced by the retrieved textual contexts.

Table~\ref{tab:online_efficiency} reports the online generation efficiency on CSL-Daily.
Compared with the Non-RAG baseline (\texttt{Pretrain$\rightarrow$SFT}), SignRAG-SFT increases the average latency from 42.69~ms/sample to 61.98~ms/sample, corresponding to an additional 19.29~ms/sample.
The throughput decreases from 23.42 to 16.13 samples/s, while the peak GPU memory increases from 23.28 to 29.03~GiB/GPU.
This corresponds to a throughput reduction of approximately 31.13\% and an additional memory consumption of 5.75~GiB/GPU.
Despite this moderate overhead, SignRAG-SFT improves BLEU-4 on CSL-Daily from 29.72 to 31.22.

\begin{table}[ht]
\centering
\caption{
\textbf{Online inference efficiency on CSL-Daily.}
Both methods use Qwen3-8B and four NVIDIA RTX 4090D GPUs under the same inference configuration.
Retrieval itself is performed offline and is therefore excluded from the deployed generation path.
}
\label{tab:online_efficiency}
\setlength{\tabcolsep}{5pt}
\begin{tabular}{lccc}
\toprule
\textbf{Method}
& \textbf{Latency}
& \textbf{Throughput}
& \textbf{Peak GPU Mem.} \\
& \textbf{(ms/sample)}
& \textbf{(samples/s)}
& \textbf{(GiB/GPU)} \\
\midrule
Non-RAG (\texttt{Pretrain$\rightarrow$SFT})
& 42.69
& 23.42
& 23.28 \\
SignRAG-SFT
& 61.98
& 16.13
& 29.03 \\
\bottomrule
\end{tabular}
\end{table}

\subsection{Offline Retrieval Efficiency on CSL-Daily}
\label{sec:csl_retrieval_efficiency}

We next evaluate the efficiency of the offline retrieval stage.
The retrieval gallery consists of the target-domain training samples, while each evaluation sample serves as a query.
Table~\ref{tab:csl_retrieval_scaling} reports the retrieval latency and throughput when progressively increasing the gallery size while keeping the same set of 256 queries.

As the gallery grows from 1K to 18.4K samples, the average retrieval latency increases from 3.49 to 23.88~ms/query.
At the same time, retrieval throughput decreases from 286.67 to 42.02 queries/s.
These results show that the retrieval cost increases with gallery size, but remains practical at the scale of CSL-Daily.

\begin{table}[ht]
\centering
\small
\caption{
\textbf{Offline retrieval efficiency on CSL-Daily with varying gallery sizes.}
All settings use the same 256 queries.
}
\label{tab:csl_retrieval_scaling}
\setlength{\tabcolsep}{4.5pt}
\renewcommand{\arraystretch}{0.95}
\begin{tabular}{@{}rcc@{}}
\toprule
\textbf{Gallery Size}
& \textbf{Latency}
& \textbf{Throughput} \\
& \textbf{(ms/query)}
& \textbf{(queries/s)} \\
\midrule
1K    & 3.49  & 286.67 \\
5K    & 7.10  & 140.80 \\
10K   & 14.65 & 69.32  \\
18.4K & 23.88 & 42.02  \\
\bottomrule
\end{tabular}
\end{table}

For the complete CSL-Daily test set, retrieving 1,176 queries from the full gallery of 18,401 training samples requires 64.70~s in total, corresponding to 55.01~ms/query.
The difference between this end-to-end retrieval latency and the 256-query micro-benchmark in Table~\ref{tab:csl_retrieval_scaling} is caused by the complete retrieval pipeline, which additionally includes data loading, batching, result aggregation, and manifest construction.
The peak active retrieval memory is 1.17~GiB per GPU, while the corresponding CPU memory usage is 10.42~GiB per rank.

Combining the one-time offline retrieval cost with SignRAG generation, the complete retrieval-and-generation evaluation on CSL-Daily requires 137.59~s for 1,176 test samples.
This corresponds to approximately 117.00~ms/sample and 0.153 GPU-hours when accounting for all four GPUs.
Since retrieval results can be pre-computed and reused, the deployed online inference latency remains the 61.98~ms/sample reported in Table~\ref{tab:online_efficiency}.

\subsection{Retrieval Scalability on OpenASL}
\label{sec:openasl_scalability}

To further evaluate scalability beyond CSL-Daily, we conduct the same retrieval analysis on OpenASL, whose training gallery contains approximately 96K samples.
Table~\ref{tab:openasl_retrieval_scaling} reports the retrieval efficiency under progressively increasing gallery sizes.

With a gallery of 95,888 samples, the retrieval latency reaches 705.85~ms/query for the 256-query benchmark, corresponding to a throughput of 1.42 queries/s.
For the complete OpenASL test set, retrieving 971 queries from the full 95,888-sample gallery requires 715.90~s in total, corresponding to 737.29~ms/query.
These results demonstrate that token-level retrieval remains feasible at the approximately 100K-gallery scale, although its cost grows substantially with gallery size.

\begin{table}[ht]
\centering
\caption{
\textbf{Offline retrieval efficiency at different gallery sizes on OpenASL.}
The same 256 queries are used for all gallery-size experiments.
}
\label{tab:openasl_retrieval_scaling}
\setlength{\tabcolsep}{6pt}
\begin{tabular}{rcc}
\toprule
\textbf{Gallery Size}
& \textbf{Latency}
& \textbf{Throughput} \\
& \textbf{(ms/query)}
& \textbf{(queries/s)} \\
\midrule
1,000
& 11.20
& 89.38 \\
5,000
& 39.98
& 25.02 \\
10,000
& 77.28
& 12.94 \\
95,888
& 705.85
& 1.42 \\
\bottomrule
\end{tabular}
\end{table}

\subsection{Training Cost}
\label{sec:training_cost}

Table~\ref{tab:compute_training} summarizes the training cost of the complete SignRAG pipeline, with CSL-News serving as the pre-training dataset and CSL-Daily as the downstream fine-tuning dataset.
The two-stage pre-training procedure requires 106.67~GPU-hours for Stage I and 277.50~GPU-hours for Stage II.
Downstream SignRAG-SFT requires 24.04~GPU-hours, while SignRAG-RFT requires an additional 5.78~GPU-hours.
Overall, the complete training pipeline requires 413.99~GPU-hours.

\begin{table}[ht]
\centering
\caption{\textbf{Training cost of SignRAG.}}
\label{tab:compute_training}
\setlength{\tabcolsep}{8pt}
\begin{tabular}{lc}
\toprule
\textbf{Component} & \textbf{Cost} \\
\midrule
Pre-training Stage I & 106.67 GPU-hours \\
Pre-training Stage II & 277.50 GPU-hours \\
SignRAG-SFT & 24.04 GPU-hours \\
SignRAG-RFT & 5.78 GPU-hours \\
\midrule
\textbf{Total training compute} & \textbf{413.99 GPU-hours} \\
\bottomrule
\end{tabular}
\end{table}

\subsection{Discussion on Million-Scale Retrieval}
\label{sec:retrieval_limitation}

The above experiments demonstrate the practical feasibility of SignRAG on CSL-Daily and OpenASL, with gallery sizes of approximately 18K and 96K, respectively. However, token-level matching becomes increasingly expensive as the gallery grows. For substantially larger corpora such as BOBSL, which contains approximately 1.2 million sentences, we do not currently report direct retrieval results. At present, BOBSL is not available to us, and a reliable evaluation would additionally require data preparation, model pretraining, feature extraction, index construction, and retrieval-quality validation. Therefore, without direct experimental evidence, we avoid extrapolating the current measurements to the million-scale setting.

A potential direction for improving scalability is \emph{coarse-to-fine retrieval}. For example, global video representations or domain information could first be used to shortlist a candidate pool through approximate nearest-neighbor search, after which the proposed fine-grained token-level matching would be applied only to the shortlisted candidates. Gallery sharding and batched retrieval could further reduce wall-clock latency.

However, coarse-to-fine retrieval introduces a non-trivial accuracy--efficiency trade-off. A coarse retrieval stage based on a single global representation may discard samples that would otherwise receive high scores under full token-level matching. Therefore, we do not claim that coarse-to-fine retrieval can fully preserve the accuracy of exhaustive SignRAG retrieval. A more comprehensive evaluation at the million-scale should jointly consider (i) retrieval latency, (ii) Top-$K$ candidate preservation, (iii) retrieval quality, and (iv) downstream SLT performance under different candidate-pool sizes.

Finally, we emphasize that the primary contribution of SignRAG is not to claim unrestricted scalability to arbitrary million-scale open-domain corpora, but rather to extend retrieval-augmented SLT from settings that rely on gloss annotations to a gloss-free formulation and to demonstrate its effectiveness on multiple downstream benchmarks, including relatively large-scale datasets such as How2Sign and OpenASL. The OpenASL results in Table~\ref{tab:openasl_retrieval_scaling} directly characterize the computational behavior of the current implementation at the largest scale evaluated in this work.

For very large datasets, the limited vocabulary of sign language may also reduce the need for global Top-$K$ retrieval. A fixed subset of the gallery may already provide useful retrieval candidates. However, this hypothesis still requires empirical validation. Million-scale retrieval therefore remains a limitation of the current work and an important direction for future study.

\section{More Cases}\label{sec:more_case}
Due to space limitations in the main paper, we provide additional qualitative examples in Table~\ref{tab:supp_qualitative_cases}. These cases offer a more comprehensive view of the translation performance of our method across diverse examples.

\begin{CJK*}{UTF8}{gbsn}
\begin{table*}[ht]
\centering
\scriptsize
\setlength{\tabcolsep}{4pt}
\renewcommand{\arraystretch}{1.06}

\begin{tabular}{@{}p{0.105\textwidth}p{0.865\textwidth}@{}}
\toprule
\textbf{Source} & \textbf{Sentence} \\
\midrule

% =========================================================
% Example 1
% =========================================================
\textbf{Reference} &
我想要找借口。 \newline
\textit{(I want to find an excuse.)} \\[1pt]

\textbf{Retrieved} &
[1] 他随口编了一个迟到的借口。
\textit{(He casually made up an excuse for being late.)} \newline
[2] 为自己找借口的人，永远不会进步。
\textit{(People who make excuses for themselves will never improve.)} \newline
[3] 我不会用出差作为任何事情的借口。
\textit{(I would not use a business trip as an excuse for anything.)} \\[1pt]

\textbf{Geo-Sign} &
你找我有什么事吗？ \newline
\textit{(Why are you looking for me?)} \\[1pt]

\textbf{SignRAG} &
找借口。 \newline
\textit{(Find an excuse.)} \\

\midrule

% =========================================================
% Example 2
% =========================================================
\textbf{Reference} &
封面设计是一种艺术设计。 \newline
\textit{(Cover design is a form of artistic design.)} \\[1pt]

\textbf{Retrieved} &
[1] 一个好的封面设计会让你脱颖而出。
\textit{(A good cover design will make you stand out.)} \newline
[2] 我姐姐是从事封面设计工作。
\textit{(My sister works in cover design.)} \newline
[3] 我姐姐是从事封面设计工作。
\textit{(My sister works in cover design.)} \\[1pt]

\textbf{Geo-Sign} &
平面设计是一种艺术与技术的结合。 \newline
\textit{(Graphic design is a combination of art and technology.)} \\[1pt]

\textbf{SignRAG} &
封面设计是一种艺术设计。 \newline
\textit{(Cover design is a form of artistic design.)} \\

\midrule

% =========================================================
% Example 3
% =========================================================
\textbf{Reference} &
这个视频拍下了你作弊的行为。 \newline
\textit{(This video recorded you cheating.)} \\[1pt]

\textbf{Retrieved} &
[1] 他对自己作弊的行为感到非常羞愧。
\textit{(He felt very ashamed of his cheating behavior.)} \newline
[2] 他对自己作弊的行为感到非常羞愧。
\textit{(He felt very ashamed of his cheating behavior.)} \newline
[3] 你的行动还不够迅速。
\textit{(Your actions are not quick enough.)} \\[1pt]

\textbf{Geo-Sign} &
他以借书为名，行偷盗之实。 \newline
\textit{(He used borrowing books as a pretext for stealing.)} \\[1pt]

\textbf{SignRAG} &
你作弊的行为是可耻的。 \newline
\textit{(Your cheating behavior is shameful.)} \\

\midrule

% =========================================================
% Example 4
% =========================================================
\textbf{Reference} &
今天水果很新鲜，买一个西瓜回家。 \newline
\textit{(The fruit is very fresh today; buy a watermelon to take home.)} \\[1pt]

\textbf{Retrieved} &
[1] 天气热，买点水果回家给爸爸吃。
\textit{(The weather is hot; buy some fruit to take home for Dad.)} \newline
[2] 这些水果真新鲜，我们买西瓜还是苹果？
\textit{(These fruits are really fresh; should we buy a watermelon or apples?)} \newline
[3] 这些水果真新鲜，我们买西瓜还是苹果？
\textit{(These fruits are really fresh; should we buy a watermelon or apples?)} \\[1pt]

\textbf{Geo-Sign} &
水果今天很新鲜，买一串香蕉回家。 \newline
\textit{(The fruit is very fresh today; buy a bunch of bananas to take home.)} \\[1pt]

\textbf{SignRAG} &
水果今天很新鲜，买一个西瓜回家。 \newline
\textit{(The fruit is very fresh today; buy a watermelon to take home.)} \\

\bottomrule
\end{tabular}

\vspace{-1.5mm}
\caption{\small
Additional qualitative comparisons on CSL-Daily.
Each example presents the reference translation, top-3 retrieved texts, and outputs of Geo-Sign and SignRAG.
English translations are shown in parentheses below the original Chinese sentences.
}
\vspace{-2mm}
\label{tab:supp_qualitative_cases}

\end{table*}
\end{CJK*}

\end{document}